\documentclass[journal]{IEEEtran}
\usepackage{amsmath,amssymb,amsfonts}
\usepackage{mathtools}
\usepackage{empheq}
\usepackage{cases}
\usepackage{color,xcolor}
\usepackage{soul}

\usepackage{graphicx}
\usepackage{stfloats}
\usepackage{placeins}
\usepackage[nocompress]{cite}
\usepackage{subfigure}
\usepackage{multirow}
\usepackage{rotating}
\usepackage{booktabs}
\usepackage{tikz}
\usepackage{acronym}
\usepackage{epstopdf}
\usepackage{pifont} % 导入pifont包以使用打钩和打叉符号
\usepackage[export]{adjustbox}
\usepackage{tabularray}
\usepackage{colortbl}
 \usepackage[colorlinks,
 linkcolor=blue,
 anchorcolor=blue,
 citecolor=blue,
 urlcolor=blue
]{hyperref}

\usepackage{amssymb}
\usepackage{array} % 引入array宏包
\usepackage{catchfilebetweentags}
\usepackage[ruled, vlined]{algorithm2e}
\usepackage{newtxmath}
\newcolumntype{Z}{>{\centering\arraybackslash}p{1.2cm}}
\newcolumntype{W}{>{\centering\arraybackslash}p{1.5cm}}

\usepackage{diagbox}
\usepackage{multicol}
\usepackage{algpseudocode}
\usepackage{algorithmicx}


\begin{document}

\title{FedSGN: Federated Sparse Grassmann Newton \\Learning for IoT Anomaly Detection}

\author{Jingjing Liu, Xiansen Ju, Xianchao Xiu, and Wanquan Liu, \IEEEmembership{Senior Member,~IEEE}
\thanks{This work was supported in part by the National Natural Science Foundation of China under Grants 62204044 and 12371306, the State Key Laboratory of Integrated Chips and Systems under Grant SKLICS-K202302, and the Shenzhen Nature Science Research Foundation Key Project under Grant ZDYJ20251211121303005. (Corresponding author: Xianchao Xiu.)}
\thanks{Jingjing Liu is with the School of Microelectronics, Shanghai Key Laboratory of Chips and Systems for Intelligent Connected Vehicle, Shanghai University, Shanghai 200444, China, and also with  the State Key Laboratory of Integrated Chips and Systems, Fudan University, Shanghai 201203, China. (email: jjliu@shu.edu.cn).}
\thanks{Xiansen Ju is with the School of Microelectronics, Shanghai Key Laboratory of Chips and Systems for Intelligent Connected Vehicle, Shanghai University, Shanghai 200444, China (email: jxs@shu.edu.cn).}
\thanks{Xianchao Xiu is with the School of Mechatronic Engineering and Automation, Shanghai University, Shanghai 200444, China (email: xcxiu@shu.edu.cn).}
\thanks{Wanquan Liu is with the School of Intelligent Systems Engineering, Sun Yat-sen University, Shenzhen 510275, China (email: liuwq63@mail.sysu.edu.cn).}}

\hypersetup{pdfauthor={Jingjing Liu, Xiansen Ju, Xianchao Xiu, Wanquan Liu}}

\maketitle

\begin{abstract}
Anomaly detection plays a critical role in securing the Internet of things (IoT). Although existing federated principal component analysis (PCA) methods can learn the subspace of normal samples without sharing raw data, they often suffer from feature redundancy that may degrade detection performance. To address this limitation, we propose a novel and effective federated sparse Grassmann Newton (FedSGN) learning method for IoT anomaly detection. Technically, our FedSGN leverages the $\ell_{2,0}$-norm constraint on the global projection matrix to eliminate redundancy and introduces an orthogonality penalty to preserve the intrinsic structure. Moreover, we develop an augmented Lagrangian solver integrated with a semismooth Newton scheme to enable parallel client updates and Grassmann projection for robustness. Numerical experiments demonstrate that the proposed method achieves favorable overall detection performance while reducing communication overhead. The source code is publicly available at \url{https://github.com/xianchaoxiu/FedSGN}.
\end{abstract}

\begin{IEEEkeywords}
Internet of things, anomaly detection, federated learning, sparse optimization, Grassmann manifold
\end{IEEEkeywords}

\section{Introduction}
\begin{figure*}[!t]
    \centering
    \includegraphics[width=0.95\textwidth]{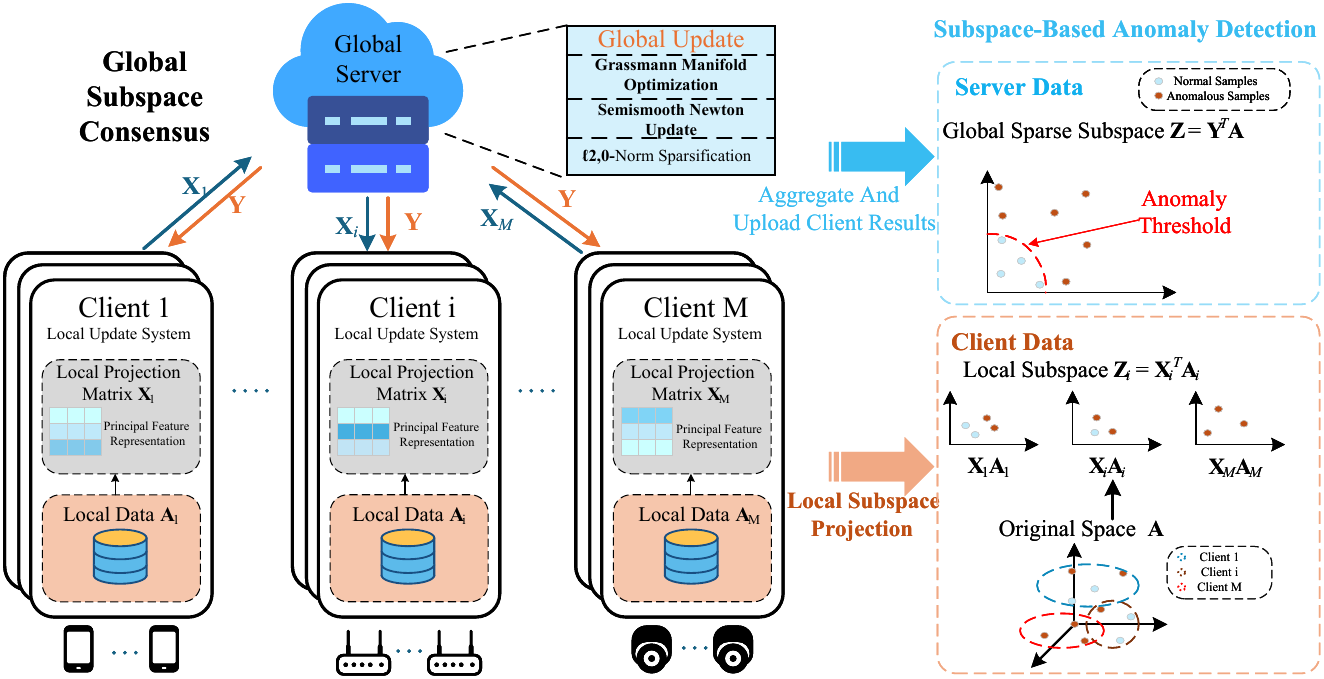}
    \caption{Framework of our proposed FedSGN. In each round, clients update their local models and send the updates to the server, which aggregates them and refines the shared sparse subspace. After training, samples with large reconstruction errors are flagged as anomalies.}
    \label{framework}
\end{figure*}

\IEEEPARstart{T}{he} Internet of things (IoT) connects different devices equipped with sensing, computing, and communication capabilities, finding extensive applications in fields such as smart manufacturing, transportation, healthcare, and urban services \cite{aouedi2024survey}. Through continuous data exchange, these systems can monitor the physical environment and respond to evolving conditions \cite{rejeb2022big}. However, this pervasive connectivity significantly expands the attack surface, leading to critical security and privacy vulnerabilities \cite{zhou2017security,tang2022computational}. Currently, anomaly detection, which aims at identifying observed data that deviates from normal behavioral patterns, has attracted intensive research attention within the IoT ecosystem. Interested readers may refer to the recent surveys  \cite{tu2025distributed,liu2024deafl,adhikari2024recent}.

Traditional IoT anomaly detection methods often collect raw data generated by devices into a central server. However, the sharing and processing of these records raise concerns regarding the use and protection of personal data \cite{kounoudes2020mapping,li2025flow}. Federated learning (FL) offers a distributed alternative in which clients retain their raw data and collaborate solely by exchanging model information \cite{mcmahan2017communication,nguyen2021federated}. Nevertheless, the heterogeneity of devices and variations in local data distributions increase the complexity of such collaboration \cite{li2020federated,lin2025fidscl}. To address these challenges in intrusion detection, regularized model aggregation is combined with adaptive differential privacy \cite{hu2025efficient}. Yet a federated detector must still balance detection capability against local computation and repeated model transmission at resource-constrained gateways.

Neural networks and deep learning have been widely applied for IoT anomaly detection \cite{abusitta2023deep,huang2025deep}. These methods learn representations of normal samples to identify deviations. An autoencoder (AE) compresses inputs through an encoder and reconstructs them through a decoder, using reconstruction errors as anomaly scores \cite{ieracitano2020novel,ruff2021unifying}. Extending this method to distributed data, federated AE (FedAE) train local models collaboratively without centralizing raw records, while FedAE2 uses a deeper architecture within the same reconstruction framework \cite{nguyen2024federated}. Beyond reconstruction learning, a bidirectional generative adversarial network (BiGAN) jointly trains an encoder, a generator, and a discriminator to learn mappings between the data and latent spaces through adversarial learning \cite{donahue2017adversarial}. Federated BiGAN (FedBiGAN) extend this framework to distributed clients, with FedBiGAN2 adopting a deeper architecture \cite{nguyen2024federated}. While these methods broaden the representation models available for federated detection, practical deployment also requires attention to model efficiency and interpretability. To reduce computational overhead, knowledge distillation has been explored for graph-based anomaly detection \cite{zhou2024reconstructed}. To further address the difficulty of explaining federated detection decisions without access to local training data, synthetic reference samples have been investigated \cite{kalakoti2026synthetic}. However, communication costs and feature interpretability remain important considerations when deploying neural network-based methods on IoT devices.

Principal component analysis (PCA) offers a lightweight way for IoT anomaly detection \cite{hoang2018pca,liu2025similarity}. As shown in \cite{carter2022fast}, by employing PCA for feature compression, the storage resource requirements for devices can be reduced. Furthermore, federated PCA enables clients to learn a subspace representation of normal samples collaboratively without the need to exchange raw data. Representative  anomaly detection methods include FedPE, which utilizes the alternating direction method of multipliers (ADMM) \cite{boyd2011distributed} in Euclidean space, and FedPG, which optimizes on an intrinsic Grassmann manifold \cite{nguyen2024federated}. More recently, FedEP  incorporates $\ell_{2,1}$-norm sparse regularization term and local models, thereby accommodating client-specific noise and enhancing system robustness \cite{xiu2026fedep}. Although such methods can reduce model complexity, they do not impose explicit constraints on the number of nonzero rows in the global projection matrix \cite{liu2026bi,li2026enhancing}. Therefore, redundant features may persist in the shared representation, which not only limits the interpretability  but also hinders the reduction of model transmission volume.

To address the above issue, we propose an effective  federated sparse Grassmann Newton learning method, abbreviated as FedSGN, for IoT anomaly detection. Different from FedEP that relies on the $\ell_{2,1}$-norm regularization term on the local model, our FedSGN imposes an $\ell_{2,0}$-norm constraint on the global projection matrix, thereby enabling explicit feature selection. To the best of our knowledge, it has not been explored in federated learning. During the optimization process, an orthogonality penalty term is introduced to preserve the subspace structure, which differs fundamentally from FedPE and FedPG that depend on singular value decomposition (SVD). Thus, the resulting projection matrix can be transmitted in the sparse form, reducing the communication overhead for the global model. The framework of our proposed FedSGN is provided in Fig. \ref{framework}.

The main contributions of the current work can be summarized in the following three aspects.
\begin{itemize}
    \item We propose FedSGN, a federated sparse subspace model designed to provide a new idea for IoT anomaly detection. By incorporating an $\ell_{2,0}$-norm constraint alongside an orthogonality penalty, our model enables global feature selection through collaborative learning among clients.
    \item We develop an efficient optimization algorithm based on the ADMM strategy, which decouples parallel client updates from global sparse updates. On the server side, a semismooth Newton method is employed to accelerate the computation of subproblems.
    \item We evaluate the proposed FedSGN using both synthetic and three real-world datasets. In addition, we visualize the feature selection achieved through sparsity and compare the model transmission volume.
\end{itemize}

The remainder of this paper is organized as follows. Section \ref{related} introduces some related works. Section \ref{method} presents our proposed model and optimization algorithm. Section \ref{experiments} reports the experimental results. Section \ref{conclusions} concludes the paper.

\section{Related Work}\label{related}

This section briefly introduces necessary notations, as well as the basics of sparse PCA and federated PCA.

\subsection{Notations}
Throughout this work, matrices are denoted by boldface uppercase letters, vectors by boldface lowercase letters, and scalars by italic letters. For $\mathbf{W}\in\mathbb{R}^{d\times p}$, its $i$th row, $j$th column, and $(i,j)$th entry are denoted by $\mathbf{w}_{i:}$, $\mathbf{w}_{:j}$, and $w_{ij}$, respectively. The inner product of two matrices is $\langle \mathbf{A},\mathbf{B}\rangle=\operatorname{Tr}(\mathbf{A}^{\top}\mathbf{B})$, where $\operatorname{Tr}(\cdot)$ denotes the trace, and $\top$ denotes the transpose. The $\ell_{2,1}$-norm and Frobenius norm are defined as $\|\mathbf{W}\|_{2,1}=\sum_{i=1}^{d}\|\mathbf{w}_{i:}\|_2$ and $\|\mathbf{W}\|_F=\sqrt{\operatorname{Tr}(\mathbf{W}^{\top}\mathbf{W})}$, respectively. Moreover, the $\ell_{2,0}$-norm counts the nonzero rows, which is defined as $\|\mathbf{W}\|_{2,0}=|\{i: \|\mathbf{w}_{i:}\|_2>0\}|$, where $|\Gamma|$ represents the cardinality of an index set $\Gamma$. In addition, the operator $\operatorname{vec}(\mathbf{W})$ stacks the columns of $\mathbf{W}$. Additional notation will be introduced where needed.

\subsection{Sparse PCA}

PCA is a classic subspace learning method that preserves the the essential information of data through a low-dimensional representation. Given a centered data matrix $\mathbf{A}\in\mathbb{R}^{d\times n}$, its reconstruction objective can be formulated as
\begin{equation}\label{eq:rw_pca}
\begin{aligned}
\min_{\mathbf{X}}\quad&
\|\mathbf{A}-\mathbf{X}\mathbf{X}^{\top}\mathbf{A}\|_F^2\\
\mathrm{s.t.}\quad&
\mathbf{X}^{\top}\mathbf{X}=\mathbf{I}_p,
\end{aligned}
\end{equation}
where $\mathbf{X}\in\mathbb{R}^{d\times p}$ is an orthonormal basis, $\mathbf{X}^{\top}\mathbf{X}=\mathbf{I}_p$ is an orthogonality constraint, which also constitutes a manifold.

However, this PCA does not explicitly constrain the number of nonzero rows in the basis matrix. In the field of intrusion detection for IoT, feature engineering can reduce the training and detection costs by selecting and constructing informative features \cite{ghubaish2024lemda}. In this regard, sparse PCA introduces sparse regularization terms into its objective \cite{zou2006sparse,journee2010generalized}. A common formulation is
\begin{equation}\label{eq:rw_row_regularization}
\begin{aligned}
\min_{\mathbf{X}}\quad&
\|\mathbf{A}-\mathbf{X}\mathbf{X}^{\top}\mathbf{A}\|_F^2+\lambda\|\mathbf{X}\|_{2,1}\\
\mathrm{s.t.}\quad&
\mathbf{X}^{\top}\mathbf{X}=\mathbf{I}_p, 
\end{aligned}
\end{equation}
where $\lambda>0$ determines the degree of sparsity in $\mathbf{X}$. If an entire row is zero, it implies that the corresponding feature has been excluded from the representation. Although problem \eqref{eq:rw_row_regularization} can promote row-wise sparsity, it cannot directly constrain the number of rows that are retained. In contrast, $\|\mathbf{X}\|_{2,0} \leq s$ provides an explicit characterization with at most $s$ nonzero rows \cite{dong2025unsupervised,zhu2025joint}. This yields the following PCA variant
\begin{equation}\label{eq:rw_row_regularization1}
\begin{aligned}
\min_{\mathbf{X}}\quad&
\|\mathbf{A}-\mathbf{X}\mathbf{X}^{\top}\mathbf{A}\|_F^2\\
\mathrm{s.t.}\quad&
\mathbf{X}^{\top}\mathbf{X}=\mathbf{I}_p, ~\|\mathbf{X}\|_{2,0}\leq s.
\end{aligned}
\end{equation}
From an optimization perspective, the combination of  $\mathbf{X}^{\top}\mathbf{X}=\mathbf{I}_p$ and $\|\mathbf{X}\|_{2,0} \leq s$ renders the problem non-convex. Consequently, it necessitates the use of appropriate solution algorithms, such as ADMM \cite{xiu2022sparsity}, manifold proximal gradient method \cite{chen2020alternating}, and Riemannian ADMM \cite{li2025riemannian}. It should be noted that they are first-order optimization algorithms, whereas this paper employs a second-order algorithm based on a semismooth Newton strategy.

\subsection{Federated PCA}

Federated PCA aims to learn a common subspace from data distributed across $M$ clients, without the need to exchange raw records \cite{grammenos2020federated,wang2023high}. Let $\mathbf{A}_i\in\mathbb{R}^{d\times n_i}$ denote the data matrix for the $i$th client, $\mathbf{X}_i$ its local basis, and $\mathbf{Y}$ the shared basis. Here, $d$ is the number of features and $n_i$ is the number of training samples at client $i$. A standard consensus-based federated PCA formulation in \cite{nguyen2024federated} is given by
\begin{equation}\label{eq:rw_federated_pca}
\begin{aligned}
\min_{\mathbf{Y},\{\mathbf{X}_i\}}\quad&
\sum_{i=1}^{M}
\|\mathbf{A}_i-\mathbf{X}_i\mathbf{X}_i^{\top}\mathbf{A}_i\|_F^2\\
\mathrm{s.t.}\quad&
\mathbf{X}_i^{\top}\mathbf{X}_i=\mathbf{I}_p,~\mathbf{X}_i=\mathbf{Y},~ i\in[M].
\end{aligned}
\end{equation}
where $[M]=\{1,\cdots,M\}$. Although the consensus constraints above can ensure that the basis is shared, they do not directly facilitate feature selection. In comparison, this paper considers federated sparse PCA which enables efficient feature selection by imposing row-wise sparse constraints on the shared basis.

\section{The Proposed Method}\label{method}

\subsection{Problem Formulation}

Motivated by the above observations, we attempt to learn a shared orthogonal subspace of dimension $p$ that contains at most $s$ nonzero rows. By integrating problem \eqref{eq:rw_row_regularization1} and problem \eqref{eq:rw_federated_pca}, it derives the following mathematical model
\begin{equation}\label{eq:our}
\begin{aligned}
\min_{\mathbf{Y},\{\mathbf{X}_i\}}\quad&
\sum_{i=1}^{M}
\|\mathbf{A}_i-\mathbf{X}_i\mathbf{X}_i^{\top}\mathbf{A}_i\|_F^2\\
\mathrm{s.t.}\quad&
\mathbf{X}_i^{\top}\mathbf{X}_i=\mathbf{I}_p,~\mathbf{X}_i=\mathbf{Y},~ i\in[M],\\
&
\|\mathbf{Y}\|_{2,0}\leq s,
\end{aligned}
\end{equation}
where $\|\mathbf{Y}\|_{2,0}$ counts the nonzero rows of $\mathbf{Y}$. The subspace dimension and feature budget satisfy $1\leq p\leq s\leq d$.

Instead, we leverage an equivalent objective function based on covariance and replace the strict orthogonality constraints with a penalty on the global matrix. This leads to the following sparse consensus problem
\begin{equation}\label{objective}
\begin{aligned}
\min_{\mathbf{Y},\{\mathbf{X}_i\}}\quad&
\sum_{i=1}^{M}f_i(\mathbf{X}_i)
+
\frac{\mu}{4}\|\mathbf{Y}^{\top}\mathbf{Y}-\mathbf{I}_p\|_F^2
\\
\mathrm{s.t.}\quad&
\mathbf{X}_i=\mathbf{Y},~i\in[M],
\\
&
\|\mathbf{Y}\|_{2,0}\leq s,
\end{aligned}
\end{equation}
where $\mu>0$ is the orthogonality penalty weight and 
\begin{equation}
f_i(\mathbf{X}_i)
=-\frac{1}{2n_i}\operatorname{Tr}
(\mathbf{X}_i^{\top}\mathbf{A}_i\mathbf{A}_i^{\top}\mathbf{X}_i).
\end{equation}

Assume that $\mathbf{Y}$ is the final global orthogonal basis obtained by solving problem \eqref{objective}. For a centered sample $\mathbf{a} \in \mathbb{R}^{d}$, the squared reconstruction residual is used as its anomaly score
\begin{equation}\label{eq:rw_anomaly_score}
r(\mathbf{a})=\|\mathbf{a}-\mathbf{Y}\mathbf{Y}^{\top}\mathbf{a}\|_2^2.
\end{equation}
A sample is classified as anomalous when its score exceeds a calibrated threshold.

\subsection{Optimization Algorithm}
Obviously, problem \eqref{objective} is a non-convex optimization problem. To this end, we resort to the ADMM algorithm for consensus optimization \cite{boyd2011distributed}, while incorporating the sparsity-constrained framework presented in \cite{qu2024distributed}. For ease of description, let
\begin{equation}
g(\mathbf{Y})
=
\frac{\mu}{4}
\|\mathbf{Y}^{\top}\mathbf{Y}-\mathbf{I}_p\|_F^2.
\end{equation}
For each consensus constraint $\mathbf{X}_i=\mathbf{Y},~i\in[M]$, introduce a Lagrange multiplier $\boldsymbol{\Lambda}_i\in\mathbb{R}^{d\times p}$ and a penalty parameter $\beta_i>0$. Consequently, the augmented Lagrangian function for problem \eqref{objective} is defined as
\begin{equation}
\mathcal{L}_{\boldsymbol{\beta}}
(\mathbf{Y},\{\mathbf{X}_i\},\{\boldsymbol{\Lambda}_i\})
=
\sum_{i=1}^{M}
\mathcal{L}_i
(\mathbf{Y},\mathbf{X}_i,\boldsymbol{\Lambda}_i)
+
g(\mathbf{Y}),
\end{equation}
where  $\boldsymbol{\beta}
=(\beta_1,\cdots,\beta_M)^{\top}$ and
\begin{equation}
\mathcal{L}_i
(\mathbf{Y},\mathbf{X}_i,\boldsymbol{\Lambda}_i)
=
f_i(\mathbf{X}_i)
-
\langle\boldsymbol{\Lambda}_i,\mathbf{X}_i-\mathbf{Y}\rangle
+
\frac{\beta_i}{2}
\|\mathbf{X}_i-\mathbf{Y}\|_F^2.
\end{equation}

Following the alternating
minimization scheme, problem \eqref{objective} can be solved by
\begin{subequations}\label{eq:update}
\begin{empheq}[left=\empheqlbrace]{align}
\mathbf{Y}^{k+1}
&=
\arg\min_{\|\mathbf{Y}\|_{2,0}\leq s}
\mathcal{L}_{\boldsymbol{\beta}}
(\mathbf{Y},\{\mathbf{X}_i^k\},\{\boldsymbol{\Lambda}_i^k\}),
\label{eq:update_a}
\\
\mathbf{X}_i^{k+1}
&=
\arg\min_{\mathbf{X}_i\in\mathbb{R}^{d\times p}}
\mathcal{L}_i
(\mathbf{Y}^{k+1},\mathbf{X}_i,\boldsymbol{\Lambda}_i^k),
\label{eq:update_b}
\\
\boldsymbol{\Lambda}_i^{k+1}
&=
\boldsymbol{\Lambda}_i^k
-
\beta_i(\mathbf{X}_i^{k+1}-\mathbf{Y}^{k+1}),
\label{eq:update_c}
\end{empheq}
\end{subequations}
where the second and third updates are performed locally by each client $i\in[M]$. Next, we will describe fast solvers for subproblem \eqref{eq:update_a} and subproblem \eqref{eq:update_b}, and the whole optimization algorithm for solving problem \eqref{objective} is summarized in Algorithm \ref{alg:FedSGN}.

\begin{algorithm*}[!t]
%\small
\setlength{\abovedisplayskip}{4pt}
\setlength{\belowdisplayskip}{4pt}
\setlength{\abovedisplayshortskip}{0pt}
\setlength{\belowdisplayshortskip}{4pt}
\caption{Optimization Algorithm for Solving Problem \eqref{objective}}
\label{alg:FedSGN}
\SetAlgoLined
\LinesNumbered
\DontPrintSemicolon
\SetNlSty{textsf}{\color{black!55}}{}
\newcommand{\FedPhase}[1]{%
    \begingroup\let\nl\relax\DontPrintSemicolon
    \strut #1\;\endgroup}
\SetKwInput{Input}{Input}
\SetKwInput{Initialize}{Initialize}
\SetKwInput{Output}{Output}
\Input{Local training matrices $\{\mathbf{A}_i\}$,
sparsity $s$, tolerance $\epsilon$, iteration limits $K_{\max}$ and $T_{\max}$, and parameters
$\alpha,\zeta>0$, $0<\rho<1$, $0<c<1/2$.}
\Initialize{$\{\mathbf{X}_i^0\}$, $\{\boldsymbol{\Lambda}_i^0\}$, and $\mathbf{Y}^0$ with $\|\mathbf{Y}^0\|_{2,0}\leq s$.}
\While{$k<K_{\max}$}{
    \FedPhase{$\vartriangleright$ \texttt{Server: global sparse update}}
    Set $t=0$ and $\mathbf{Y}^{k,0}=\mathbf{Y}^k$.\;
    \While{$t<T_{\max}$}{
        Select the support
        $\Gamma^{k,t}\in\Gamma_s(\mathbf{Y}^{k,t},\alpha)$.\;
        \lIf{$\|H(\mathbf{Y}^{k,t},\Gamma^{k,t})\|_F\leq\epsilon$}{break}
        Write $\mathbf{Y}=\mathbf{Y}^{k,t}$ and $\Gamma=\Gamma^{k,t}$.
        Compute $\mathbf{D}$ using \eqref{eq:reduced_newton} and
        \eqref{eq:off_support_direction}, and set $\mathbf{G}=(\nabla h(\mathbf{Y}))_{\Gamma}$.\;
        Accept the Newton direction if the system is solvable and
        \[
        \langle \mathbf{G},\mathbf{D}_{\Gamma}\rangle+\zeta\|\mathbf{D}_{\Gamma}\|_F^2
        \leq\frac{\|\mathbf{Y}_{\Gamma^c}\|_F^2}{4\alpha}.
        \]
        Otherwise, set $\mathbf{D}_{\Gamma}=-\mathbf{G}$ and
        $\mathbf{D}_{\Gamma^c}=-\mathbf{Y}_{\Gamma^c}$.\;
        Choose the step-size $\eta$ according to
        \[
        h(\mathbf{Y}+\eta \mathbf{D})\leq h(\mathbf{Y})+c\eta\langle\nabla h(\mathbf{Y}),\mathbf{D}\rangle.
        \]
        Update $\mathbf{Y}^{k,t+1}$ by \eqref{eq:Y_update}. \;
        Set $t=t+1$\;
    }
    Set $\mathbf{Y}^{k+1}=\mathbf{Y}^{k,t}$ and broadcast it.\;
    \FedPhase{$\vartriangleright$ \texttt{Clients: parallel local updates}}
    \ForEach{client $i\in[M]$ in parallel}{
        Update $\mathbf{X}_i^{k+1}$ by \eqref{eq:X_update}.\;
        Update $\boldsymbol{\Lambda}_i^{k+1}$ by \eqref{eq:dual_update}.\;
    }
    Set $k=k+1$\;
    \FedPhase{$\vartriangleright$ \texttt{Grassmann projection}}
    Compute the QR factorization
    $\mathbf{Y}^k=\mathbf{Q}\mathbf{R}$, and set $\mathbf{Y}^k\leftarrow\mathbf{Q}$ as the Grassmann subspace representation.\;
}
\Output{Global projection matrix $\mathbf{Y}$.}

\end{algorithm*}

\subsubsection{Update $\mathbf{Y}$}
When $\{\mathbf{X}_i^k\}$ and $\{\boldsymbol{\Lambda}_i^k\}$ are fixed, the global update  in subproblem \eqref{eq:update_a} can be rewritten as
\begin{equation}
\begin{aligned}
\min_{\mathbf{Y}}\quad
&h(\mathbf{Y})
=
\frac{\mu}{4}
\|\mathbf{Y}^{\top}\mathbf{Y}-\mathbf{I}_p\|_F^2\\
&+
\sum_{i=1}^{M}
\frac{\beta_i}{2}
\|
\mathbf{X}_i^k-\mathbf{Y}-\boldsymbol{\Lambda}_i^k/\beta_i
\|_F^2
\\
\mathrm{s.t.}\quad&\|\mathbf{Y}\|_{2,0}\leq s .
\end{aligned}
\end{equation}

Actually, the above problem does not admit a closed-form solution. The gradient of
$h(\mathbf{Y})$ is given by
\begin{equation}
\nabla h(\mathbf{Y})
=
\mu \mathbf{Y} (\mathbf{Y}^{\top}\mathbf{Y}-\mathbf{I}_p)
-
\sum_{i=1}^{M}
\beta_i
(
\mathbf{X}_i^k-\mathbf{Y}-\boldsymbol{\Lambda}_i^k/\beta_i
).
\label{eq:Y_gradient}
\end{equation}
Furthermore, the Hessian matrix  is 
\begin{equation}
\nabla^2 h(\mathbf{Y})
=
\mu\Phi(\mathbf{Y})
+
(
\sum_{i=1}^{M}\beta_i-\mu)\mathbf{I}_{dp},
\label{eq:Y_hessian}
\end{equation}
where
\begin{equation}
\Phi(\mathbf{Y})
=
\mathbf{I}_p\otimes (\mathbf{Y}\mathbf{Y}^{\top})
+
(\mathbf{Y}^{\top}\mathbf{Y})\otimes \mathbf{I}_d
+
(\mathbf{Y}^{\top}\otimes \mathbf{Y})\mathbf{P},
\label{eq:Psi}
\end{equation}
where $\otimes$ denotes the Kronecker product, and
$\mathbf{P}\in\mathbb{R}^{dp\times dp}$ is the permutation matrix satisfying
\begin{equation}
\operatorname{vec}(\mathbf{D}^{\top})
=
\mathbf{P}\operatorname{vec}(\mathbf{D}),
\end{equation}
where $\mathbf{D}\in\mathbb{R}^{d\times p}$ is an arbitrary matrix.

Let $\Gamma_s(\mathbf{Y},\alpha)$ denote the collection of index sets
corresponding to the $s$ largest row norms of
$\mathbf{Y}-\alpha\nabla h(\mathbf{Y})$, where $\alpha>0$ is a step-size parameter.
For any given support set
$\Gamma\in\Gamma_s(\mathbf{Y},\alpha)$, the sparse projection
condition can be expressed as
\begin{equation}
\mathbf{Y}
=
\Pi_{s}
(
\mathbf{Y}-\alpha\nabla h(\mathbf{Y})
),
\label{eq:projection_fixed}
\end{equation}
where $\Pi_{s}(\cdot)$ denotes the row-wise hard thresholding
operator, which retains the $s$ rows with the largest $\ell_2$-norm values
and sets the remaining rows to zero. To characterize the stationarity
condition of the sparse projection, we define the projection residual
as
\begin{equation}
\mathbf{O}
=
\mathbf{Y}-
\Pi_{s}
(
\mathbf{Y}-\alpha\nabla h(\mathbf{Y})
).
\label{eq:projection_residual}
\end{equation}

By rearranging the rows of $\mathbf{Y}$ according to the support set
$\Gamma$ and its complementary set $\Gamma^{c}$, the
projection residual can be rewritten as
\begin{equation}
\begin{aligned}
\mathbf{O}
&=
\begin{bmatrix}
\mathbf{Y}_{\Gamma}
\\
\mathbf{Y}_{\Gamma^{c}}
\end{bmatrix}
-
\begin{bmatrix}
\mathbf{Y}_{\Gamma}
-\alpha
(\nabla h(\mathbf{Y}))_{\Gamma}
\\
\mathbf{0}
\end{bmatrix}\\
&=
\begin{bmatrix}
\alpha(\nabla h(\mathbf{Y}))_{\Gamma}
\\
\mathbf{Y}_{\Gamma^{c}}
\end{bmatrix},
\end{aligned}
\label{eq:residual_expansion}
\end{equation}
and at the stationary point, it is equivalent to
\begin{equation}
(\nabla h(\mathbf{Y}))_{\Gamma}
=
\mathbf{0},~
\mathbf{Y}_{\Gamma^{c}}
=
\mathbf{0}.
\label{eq:sparse_stationary}
\end{equation}
Thus, the stationary condition of the $\mathbf{Y}$-subproblem
can be formulated as the following nonlinear equation
\begin{equation}
H(\mathbf{Y},\Gamma)
=
\begin{bmatrix}
(\nabla h(\mathbf{Y}))_{\Gamma}
\\
\mathbf{Y}_{\Gamma^{c}}
\end{bmatrix}
=
\mathbf{0}.
\label{eq:H_equation}
\end{equation}
Once the support set $\Gamma$ is determined,
$H(\mathbf{Y},\Gamma)$ is differentiable with respect to $\mathbf{Y}$. 

Below, we discuss how to solve  \eqref{eq:H_equation} for seeking a stationary point on the selected support set.
For a fixed outer iteration $k$, let 
$\mathbf{Y}^t=\mathbf{Y}^{k,t}$ for the inner Newton iterate, with $\mathbf{Y}^0=\mathbf{Y}^k$.
At inner iteration $t$, choose $\Gamma^t\in\Gamma_s(\mathbf{Y}^t,\alpha)$.
In this paper, we utilize a semismooth Newton method \cite{xiao2018regularized} with hard thresholding \cite{zhou2021global} to compute the search direction $\mathbf{D}^t$ by solving
\begin{equation}
\nabla H(\mathbf{Y}^t,\Gamma^t)
\operatorname{vec}(\mathbf{D}^t)
=
-\operatorname{vec}
(H(\mathbf{Y}^t,\Gamma^t)
),
\label{eq:newton_direction}
\end{equation}
where $\nabla H(\mathbf{Y}^t,\Gamma^t)$ represents its Jacobian matrix evaluated at the current iterate $\mathbf{Y}^t$.

To index the Hessian of $h$ by vectorized coordinates, define
\begin{equation}
\mathbb{I}(\Gamma)=\{i+(j-1)d:  i\in\Gamma, j=1,\cdots,p\}.
\end{equation}
Here, $\Gamma$ indexes feature rows,
$\mathbb{I}_t=\mathbb{I}(\Gamma^t)$, and
$\mathbb{I}_t^c=\mathbb{I}(\Gamma^{t,c})$.
Grouping coordinates in the order of
$\operatorname{vec}(\mathbf{Y}_{\Gamma^t})$ and
$\operatorname{vec}(\mathbf{Y}_{\Gamma^{t,c}})$, respectively, gives
\begin{equation}
\nabla H(\mathbf{Y}^t,\Gamma^t)
=
\begin{bmatrix}
(\nabla^2 h(\mathbf{Y}^t))_{\mathbb{I}_t\mathbb{I}_t}
&
(\nabla^2 h(\mathbf{Y}^t))_{\mathbb{I}_t\mathbb{I}_t^c}
\\
\mathbf{0}
&
\mathbf{I}_{(d-s)p}
\end{bmatrix},
\label{eq:Jacobian}
\end{equation}
where
$(\nabla^2 h(\mathbf{Y}^t))_{\mathbb{I}_t\mathbb{I}_t}$
denotes the Hessian submatrix corresponding to the selected feature
rows, and
$\mathbf{I}_{(d-s)p}$ is an identity matrix associated with the unselected
feature dimensions.

Due to the block structure of the Jacobian matrix, the Newton
direction can be efficiently obtained by solving a reduced linear
system. Furthermore, the unselected rows of the current iterate need not be zero when the support changes. Therefore, the Newton equation in
\eqref{eq:newton_direction} reduces to
\begin{equation}
\begin{aligned}
&(\nabla^2 h(\mathbf{Y}^t))_{\mathbb{I}_t\mathbb{I}_t}
\operatorname{vec}(\mathbf{D}^t_{\Gamma^t})\\
&=(\nabla^2 h(\mathbf{Y}^t))_{\mathbb{I}_t\mathbb{I}_t^c}
\operatorname{vec}(\mathbf{Y}^t_{\Gamma^{t,c}})
-\operatorname{vec}((\nabla h(\mathbf{Y}^t))_{\Gamma^t}).
\end{aligned}
\label{eq:reduced_newton}
\end{equation}
The remaining direction rows are given by
\begin{equation}
\mathbf{D}^t_{\Gamma^{t,c}}=-\mathbf{Y}^t_{\Gamma^{t,c}}.
\label{eq:off_support_direction}
\end{equation}
The reduced Hessian need not be nonsingular. If the Newton system
cannot be solved, the negative gradient on the selected rows is used
as a fallback direction, while \eqref{eq:off_support_direction} is retained.

To preserve row-wise sparse constraints, update $\mathbf{Y}^{t+1}=\mathbf{Y}^t(\eta_t)$, where,
after ordering the rows by $\Gamma^t$ and its complement via
\begin{equation}
\mathbf{Y}^t(\eta)=
\begin{bmatrix}
\mathbf{Y}^t_{\Gamma^t}+\eta \mathbf{D}^t_{\Gamma^t}\\
\mathbf{0}
\end{bmatrix}.
\label{eq:Y_update}
\end{equation}
Here,  the step-size $\eta_t\in(0,1]$ is chosen by the Armijo line search
specified in Algorithm \ref{alg:FedSGN}.

\subsubsection{Update $\mathbf{X}_i$}

Given the updated global subspace matrix $\mathbf{Y}^{k+1}$, each client
computes its local variable $\mathbf{X}_i$ by solving
\begin{equation}
\mathbf{X}_i^{k+1}
=
\arg\min_{\mathbf{X}_i}~
\mathcal{L}_i
(\mathbf{Y}^{k+1},\mathbf{X}_i,\boldsymbol{\Lambda}_i^k),
\label{eq:X_subproblem}
\end{equation}
which is equivalent to
\begin{equation}
\mathbf{X}_i^{k+1}
=
\arg\min_{\mathbf{X}_i}~f_i(\mathbf{X}_i)+
\frac{\beta_i}{2}
\|\mathbf{X}_i-\mathbf{Y}^{k+1} - \boldsymbol{\Lambda}_i^{k}/\beta_i\|_F^2.
\label{eq:X_subproblem}
\end{equation}

Taking the derivative with respect to $\mathbf{X}_i$ and setting it to zero yields
\begin{equation}
\nabla f_i(\mathbf{X}_i)
-
\boldsymbol{\Lambda}_i^k
+
\beta_i(\mathbf{X}_i-\mathbf{Y}^{k+1})
=
\mathbf{0}.
\label{eq:X_gradient}
\end{equation}
Recall that
\begin{equation}
\nabla f_i(\mathbf{X}_i)
=-\frac{1}{n_i}\mathbf{A}_i\mathbf{A}_i^{\top}\mathbf{X}_i,
\end{equation}
then substituting it into \eqref{eq:X_gradient}, it holds 
\begin{equation}
-\frac{1}{n_i}\mathbf{A}_i\mathbf{A}_i^{\top}\mathbf{X}_i
-
\boldsymbol{\Lambda}_i^k
+
\beta_i(\mathbf{X}_i-\mathbf{Y}^{k+1})
=
\mathbf{0}.
\end{equation}
Rearranging the above equation gives the closed-form solution
\begin{equation}
\mathbf{X}_i^{k+1}
=
(
\beta_i\mathbf{I}_d-\frac{1}{n_i}\mathbf{A}_i\mathbf{A}_i^{\top}
)^{-1}
(
\beta_i\mathbf{Y}^{k+1}
+
\boldsymbol{\Lambda}_i^k).
\label{eq:X_update}
\end{equation}

\subsubsection{Update $\boldsymbol{\Lambda}_i$}
After obtaining $\mathbf{Y}^{k+1}$ and $\{\mathbf{X}_i^{k+1}\}$, the dual variables are calculated as
\begin{equation}
\boldsymbol{\Lambda}_i^{k+1}
=
\boldsymbol{\Lambda}_i^k
-
\beta_i
(\mathbf{X}_i^{k+1}-\mathbf{Y}^{k+1}).
\label{eq:dual_update}
\end{equation}

\subsection{Complexity Analysis}
In each communication round, the server performs at most $T_{max}$ sparse Newton steps, each costing $O((sp)^3 + dp^2 + d\log d)$, followed by a Grassmann projection costing $O(dp^2)$. The $M$ clients perform their local updates in parallel, with a total computational cost of $O(Md^3)$. The server sends at most $sp$ coefficients and $s$ row indices to each client, while each client uploads $O(dp)$ values. Thus, row-wise sparse constraints reduce the broadcast volume, and clients exchange model variables without sharing raw data.

\begin{table}[!t]
    \centering
    \caption{Statistics of the real-world datasets}
    \label{tab:datasets}
    \renewcommand\arraystretch{1.15}
    \begin{tabular}{lcccc}
        \toprule
        Datasets & Features & Training & Testing & Classes \\
        \midrule
        UNSW-NB15 & 39 & 65,000 & 65,332 & 10 \\
        NSL-KDD & 34 & 125,973 & 22,544 & 5 \\
        TON-IoT & 49 & 114,956 & 66,557 & 10 \\
        \bottomrule
    \end{tabular}
\end{table}

\begin{figure*}[!tbp]
	\centering
	\subfigure[FedPG]{%
		\includegraphics[width=0.30\textwidth]{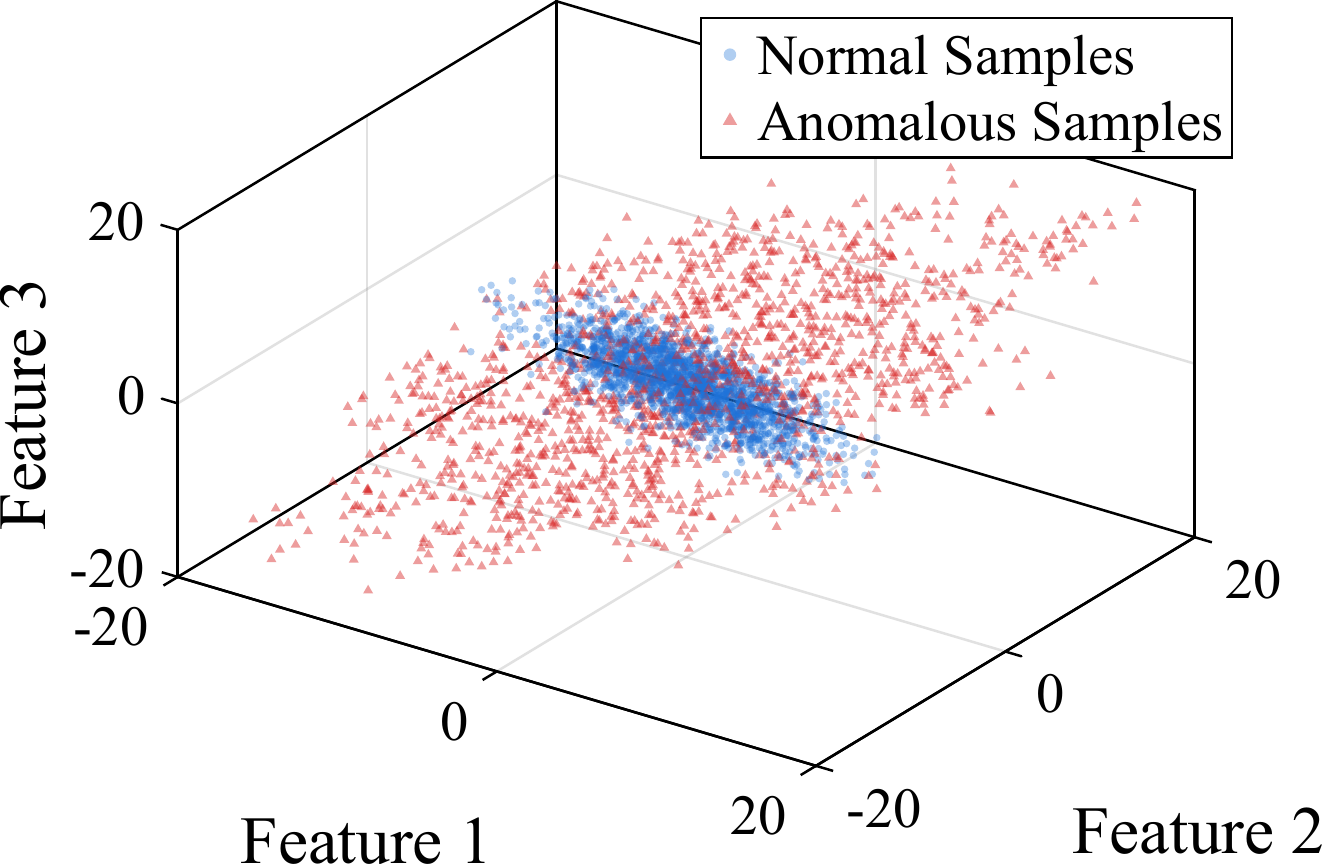}%
		\label{fig:FedPG_gap}%
	}\qquad
	\subfigure[FedEP]{%
		\includegraphics[width=0.30\textwidth]{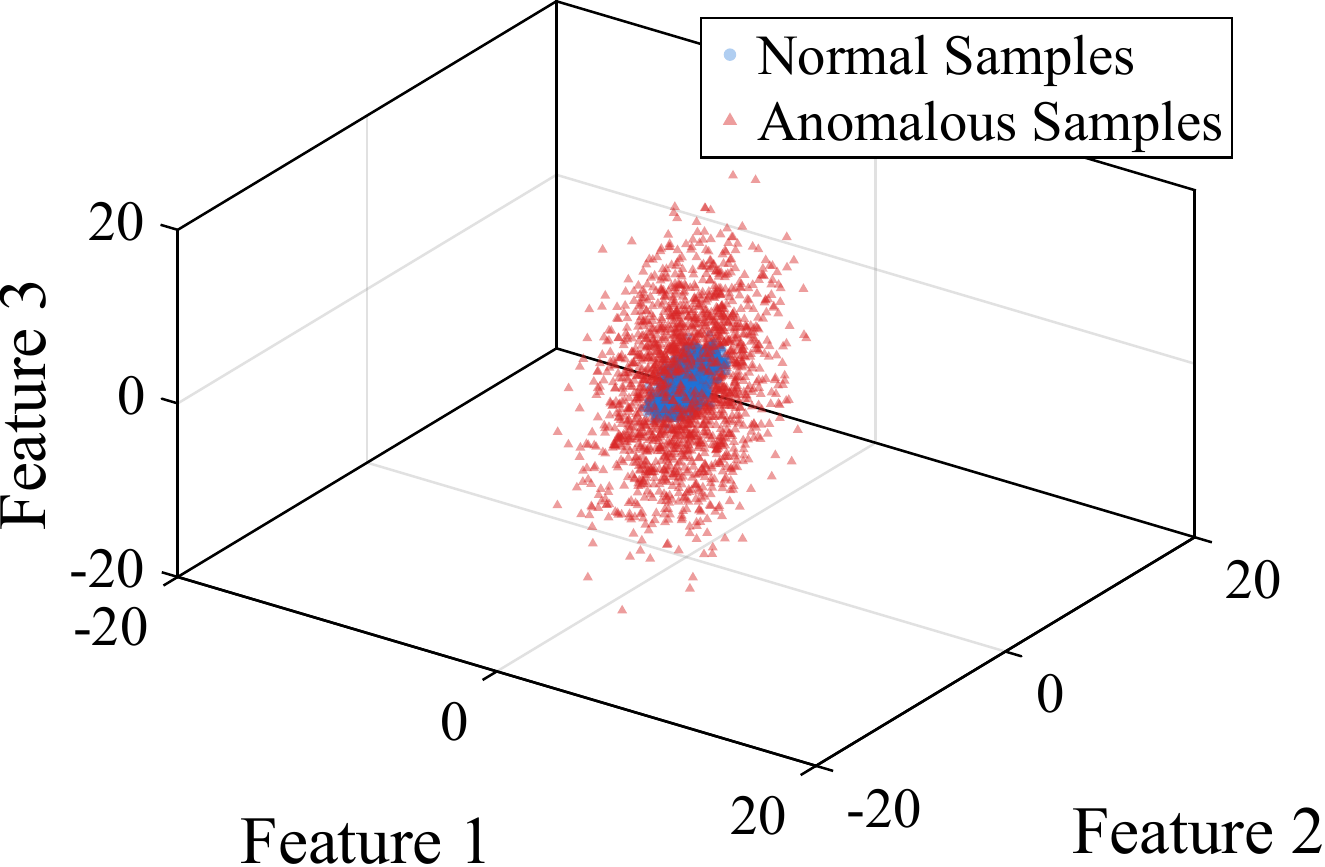}%
		\label{fig:FedEP_gap}%
	}\qquad
	\subfigure[FedSGN]{%
		\includegraphics[width=0.30\textwidth]{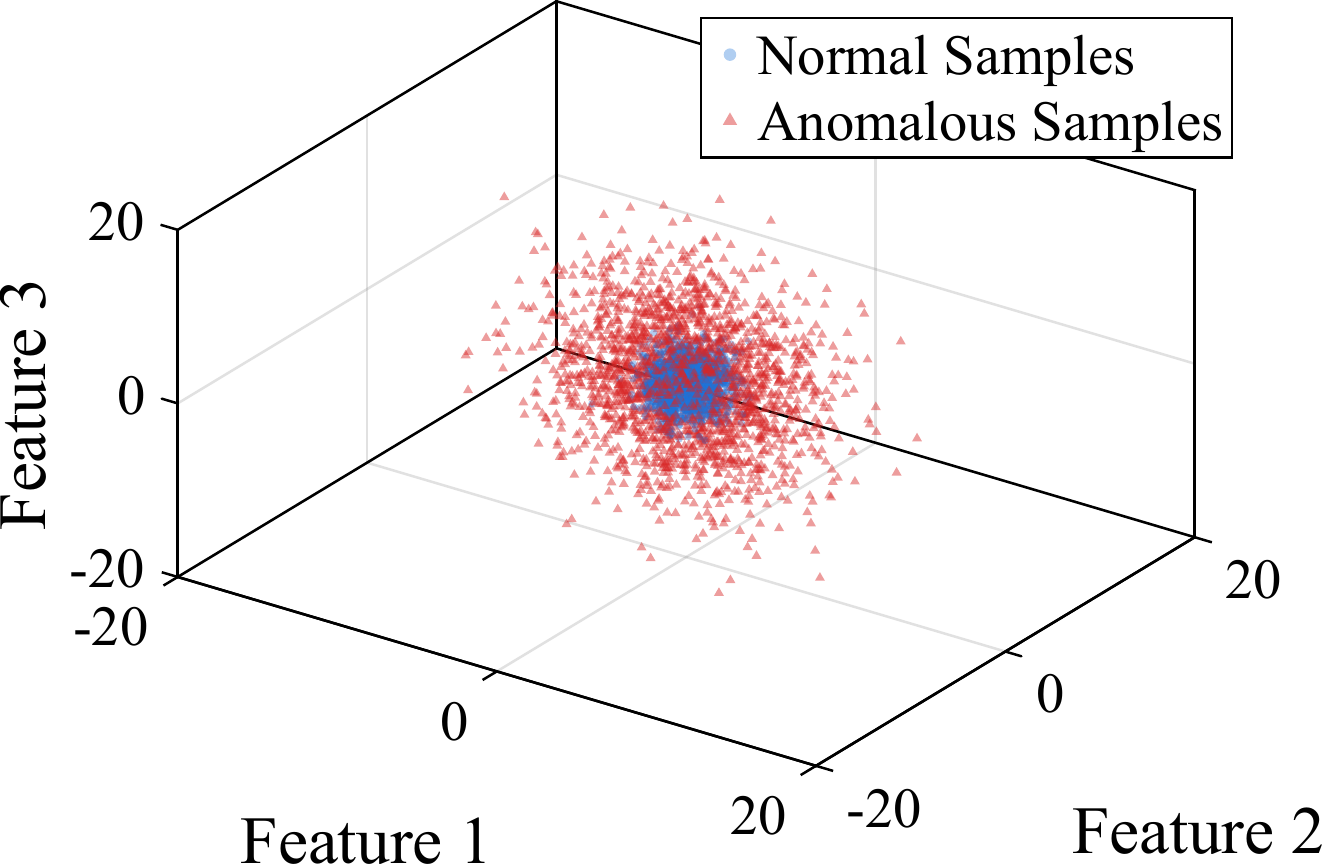}%
		\label{fig:FedSGN_gap}%
	}
%	\vspace{-2mm}
	\caption{Three-dimensional visualization of reconstructed normal and anomalous samples.}
	\label{fig:synthetic_visualization}%
\end{figure*}

\begin{figure*}[!tbp]
	\centering
	\subfigure[AUC]{%
		\includegraphics[width=0.30\textwidth]{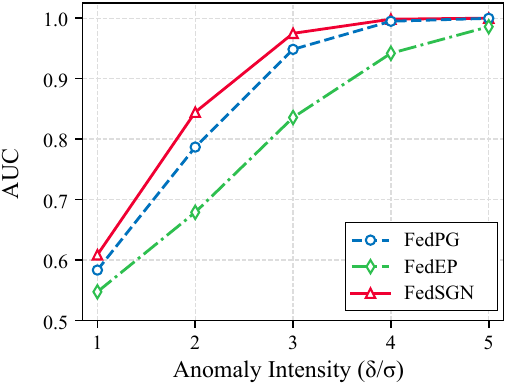}%
		\label{fig:aug}%
	}\qquad
	\subfigure[REC]{%
		\includegraphics[width=0.30\textwidth]{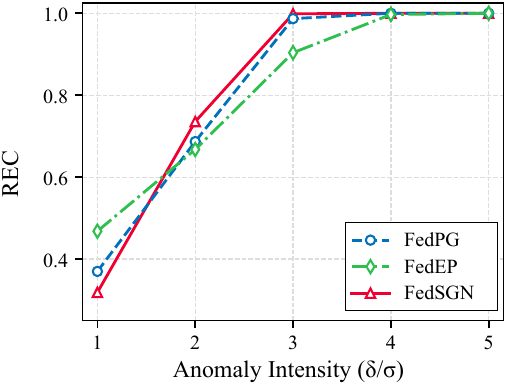}%
		\label{fig:rec}%
	}\qquad
	\subfigure[F1]{%
		\includegraphics[width=0.30\textwidth]{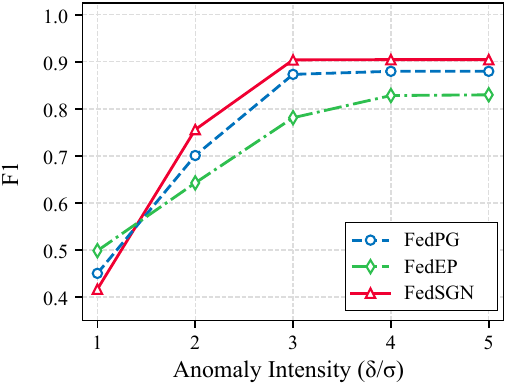}%
		\label{fig:f1}%
	}
%	\vspace{-2mm}
	\caption{Detection performance under different anomaly intensities for the synthetic datasets.}
	\label{fig:inte}%
\end{figure*}

\section{Numerical Experiments}\label{experiments}

This section begins by describing the experimental setups, then presents the experimental results on both synthetic and real-world datasets, and finally conducts ablation studies and provides further discussion.

\subsection{Experimental Settings}
\subsubsection{Compared Methods}
To demonstrate the superiority of our proposed FedSGN, the following seven state-of-the-art federated anomaly detection methods are considered.
\begin{itemize}
    \item \textbf{FedAE} \cite{hinton2006reducing}: A federated autoencoder that detects anomalies from reconstruction errors, which provides a nonlinear reconstruction baseline.
    \item \textbf{FedAE2} \cite{nguyen2024federated}: A deeper variant of FedAE that also uses reconstruction errors for anomaly detection.
    \item \textbf{FedBiGAN} \cite{donahue2017adversarial}: A federated bidirectional generative adversarial network that uses adversarial training to learn a representation of normal data.
    \item \textbf{FedBiGAN2} \cite{nguyen2024federated}: A deeper variant of FedBiGAN that retains its bidirectional adversarial training scheme.
    \item \textbf{FedPE} \cite{nguyen2024federated}: A federated PCA method that uses ADMM in Euclidean space, which is a subspace baseline without Grassmann optimization.
    \item \textbf{FedPG}  \cite{nguyen2024federated}: A federated PCA method that optimizes directly on the Grassmann manifold, which represents a geometric baseline.
    \item \textbf{FedEP} \cite{xiu2026fedep}: A personalized federated PCA method that promotes element-wise sparsity in local representations and row-wise  regularization in local models.
\end{itemize}

\subsubsection{Dataset Description}
The experiments involve both synthetic and real-world data. Here, the synthetic dataset contains 50-dimensional normal and anomalous samples generated from controlled distributions, with the anomaly intensity governed by the normalized mean shift $\delta/\sigma \in \{1, 2, 3, 4, 5\}$. Table \ref{tab:datasets} lists the selected real-world datasets, including 
\begin{itemize}
    \item \textbf{UNSW-NB15} \cite{moustafa2015unsw} combines contemporary normal samples with nine categories of attacks. After preprocessing, 39 numerical features are retained.
    \item \textbf{NSL-KDD} \cite{tavallaee2009detailed} is a refined version of the KDD'99 dataset with redundant records reduced. Its labels cover normal samples as well as four attack categories: DoS, Probe, R2L, and U2R.
    \item \textbf{TON-IoT} \cite{booij2021ton_iot} comprises heterogeneous telemetry and network attacks, including DDoS, DoS, scanning, ransomware, and injection activities.
\end{itemize}

\subsubsection{Evaluation Metrics}

According to \cite{xiu2026fedep}, five standard metrics are utilized to evaluate the performance of the compared methods. Let $\textrm{TP}$, $\textrm{TN}$, $\textrm{FP}$, and $\textrm{FN}$ represent true positives, true negatives, false positives, and false negatives, respectively.
\begin{itemize}
    \item \textbf{Accuracy ($\textrm{ACC}$)}: The overall proportion of correctly classified records, defined as
    \begin{equation*}
        \textrm{ACC} = \frac{\textrm{TP} + \textrm{TN}}{\textrm{TP} + \textrm{TN} + \textrm{FP} + \textrm{FN}}.
    \end{equation*}
    \item \textbf{Precision ($\textrm{PRE}$)}: The ratio of correctly identified attacks to all predicted attack records, defined as
    \begin{equation*}
        \textrm{PRE} = \frac{\textrm{TP}}{\textrm{TP} + \textrm{FP}}.
    \end{equation*}
    \item \textbf{Recall ($\textrm{REC}$)}: The ratio of correctly detected attacks to all actual attack records, defined as
    \begin{equation*}
        \textrm{REC} = \frac{\textrm{TP}}{\textrm{TP} + \textrm{FN}}.
    \end{equation*}
    \item \textbf{False negative rate ($\textrm{FNR}$)}: The fraction of actual anomalies that are incorrectly classified as normal, defined as
    \begin{equation*}
        \textrm{FNR} = \frac{\textrm{FN}}{\textrm{TP} + \textrm{FN}}.
    \end{equation*}
    \item \textbf{F1 score ($\textrm{F1}$)}: The harmonic mean of PRE and REC, providing a balanced measure of model performance, defined as
    \begin{equation*}
        \textrm{F1}  = \frac{2\textrm{TP}}{2\textrm{TP} + \textrm{FP} + \textrm{FN}}.
    \end{equation*}
\end{itemize}
It should be noted that higher values for ACC, PRE, REC, and F1, along with a lower FNR value, indicate better IoT anomaly  detection performance. Furthermore, receiver operating characteristic (ROC) curves and the area under the ROC curve (AUC) are also reported to evaluate discrimination across decision thresholds.

\subsubsection{Implementation Issues}
For real-world datasets, categorical fields are encoded, and numerical features are standardized using statistics fitted to the normal training data. Normal training records are distributed across clients, while raw samples remain local. For the proposed FedSGN, the subspace dimension is set to $p=10$, the row support size to $s=15$, and the number of communication rounds to 20. In addition, all compared methods utilize the same data preprocessing and partitioning, and specific setups are detailed in the corresponding subsections.

\subsection{Synthetic Results}

\begin{table*}[!t]
    \centering
    \caption{Detection performance (\%) on different datasets, where the best results are marked in bold.}
    \label{tab:detection_all}
    \renewcommand\arraystretch{1.3}
    \setlength{\tabcolsep}{4pt} 
    \begin{tabular}{lcccccccccccccccc} \toprule
        \multirow{2}{*}{Methods}  & \multicolumn{5}{c}{UNSW-NB15} & \multicolumn{5}{c}{NSL-KDD} & \multicolumn{5}{c}{TON-IoT}\\
        \cmidrule(lr){2-6} \cmidrule(lr){7-11} \cmidrule(lr){12-16}
        & ACC $\uparrow$ & PRE $\uparrow$ & REC $\uparrow$ & FNR $\downarrow$  & F1 $\uparrow$
        & ACC $\uparrow$ & PRE $\uparrow$ & REC $\uparrow$ & FNR $\downarrow$  & F1 $\uparrow$
            & ACC $\uparrow$ & PRE $\uparrow$ & REC $\uparrow$ & FNR $\downarrow$  & F1 $\uparrow$\\
        \midrule
        FedAE 
        & 80.88 & 80.66 & 95.27 & 4.73 & 87.36
        & 83.74 & 86.84 & 84.19 & 15.81 & 85.50 
        & 87.89 & 93.52 & 91.36 & 8.64 & 92.43\\
        FedAE2 
        & 82.80 & 83.45 & 93.80 & 6.20 & 88.32
        & 83.35 & 86.90 & 83.32 & 16.68 & 85.07 
        & 89.81 & 94.63 & 92.34 & 7.66 & 93.47\\
        FedBiGAN 
        & 81.49 & 83.04 & 91.63 & 8.37 & 87.12
        & 83.36 & 89.93 & 79.69 & 20.31 & 84.50 
        & 88.83 & 92.42 & 92.42 & 7.57 & 92.42\\
        FedBiGAN2 
        & 81.59 & 81.98 & 94.16 & 5.84 & 87.65
        & 82.01 & 83.71 & \textbf{84.92} & \textbf{15.08} & 84.31 
        & 89.53 & \textbf{95.82} & 91.93 & 8.07 & 93.83\\  \midrule
        FedPE 
        & 82.15 & 81.06 & 95.14 & 4.86 & 87.54
        & 81.91 & 85.62 & 81.99 & 18.01 & 83.77 
        & 87.15 & 89.17 & 94.94 & 5.06 & 91.96\\
        FedPG 
        & 81.95 & 82.82 & 93.36 & 6.64 & 87.77
        & 81.61 & 83.25 & 84.75 & 15.24 & 83.99 
        & 88.89 & 92.73 & 94.32 & 5.68 & 93.52\\
        FedEP 
        & 83.31 & 82.17 & \textbf{97.00} & \textbf{3.00} & 88.97
        & 84.24 & 89.66 & 81.75 & 18.25 & 85.52 
        & 90.48 & 92.61 & \textbf{96.65} & \textbf{3.35} & 94.59\\ \midrule
      \rowcolor{gray!20}  FedSGN (Ours)
        
        & \textbf{87.92} & \textbf{87.66} & 96.13 & 3.87 & \textbf{91.70}
        
        & \textbf{85.55} & \textbf{91.18} & 82.61 & 17.39 & \textbf{86.68}
        & \textbf{91.33} & 93.39 & 96.64 & 3.36 & \textbf{94.99}\\
        \bottomrule
    \end{tabular}
\end{table*}

In the experiments, we construct a synthetic dataset in which each sample consists of 50 features, with both normal and anomalous records generated. FedPG and FedEP are selected as federated subspace baselines for comparison and the results are shown in Fig. \ref{fig:synthetic_visualization}. Here, blue points represent normal samples and red points represent anomalous ones. It can be observed that the representations learned by different methods vary. However, the degree of compactness or overlap exhibited within this three-dimensional space does not directly reflect performance in anomaly detection. This is because the detector actually makes its decisions based on reconstruction residuals in the original feature space, rather than on the distances shown in the visualization.

Furthermore, to examine performance under varying degrees of anomalies, we keep the normal samples unchanged while constructing anomalous samples with different anomaly intensities. Anomaly intensity is represented by the standardized mean displacement $\delta/\sigma$. Here, $\delta$ denotes the displacement magnitude relative to the normal distribution, and $\sigma$ denotes the standard deviation of normal samples.
As shown in Fig. \ref{fig:inte}, the AUC, REC, and F1 metrics generally improve as the anomaly intensity increases. This indicates that the greater the deviation from the normal distribution, the easier it is to identify the anomalous samples. Across the evaluated range of anomaly intensities, FedSGN consistently achieves higher AUC values than the other two methods. At higher anomaly intensities, the REC values for all three methods tend to converge, yet FedSGN maintains its advantage in terms of F1. Overall, these results indicate favorable anomaly discrimination and detection performance of our FedSGN.

\subsection{Real-World Results}

In this section, FedSGN is evaluated using the UNSW-NB15, NSL-KDD, and TON-IoT datasets. We first compare detection performance, followed by convergence behavior and sparse feature selection.

\subsubsection{Detection Performance}
As presented in Table \ref{tab:detection_all}, our FedSGN achieves the highest ACC and F1 on all three datasets. Specifically, on the UNSW-NB15 dataset, it obtains an ACC of 87.92\% and an F1 of 91.70\%, both exceeding those of the compared methods. Moreover, FedEP achieves higher REC on the UNSW-NB15 dataset and slightly higher REC on the TON-IoT dataset, while FedBiGAN2 obtains higher PRE on the TON-IoT dataset. Nevertheless, the lead held by FedSGN in terms of ACC and F1 indicates better overall classification performance and a favorable tradeoff between PRE and REC. These findings align with FedSGN, which involves collaboratively learning a shared normal subspace while explicitly filtering for key feature rows.

\begin{figure*}[!tbp]
	\centering
	\subfigure[UNSW-NB15]{%
		\includegraphics[width=0.30\textwidth]{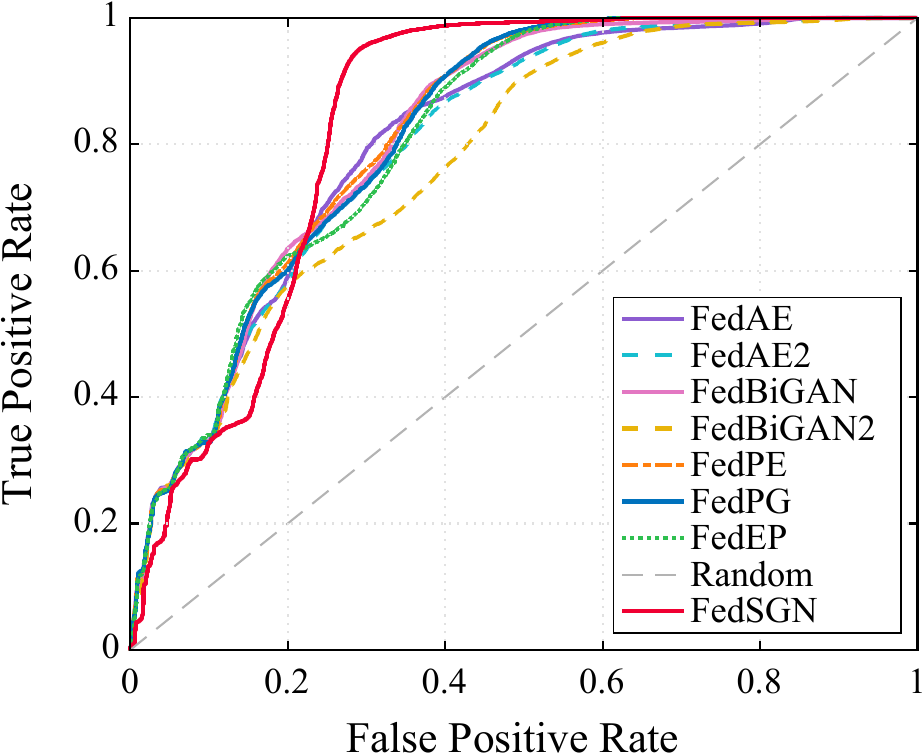}%
		\label{fig:roc_unsw}%
	}\qquad
	\subfigure[NSL-KDD]{%
		\includegraphics[width=0.30\textwidth]{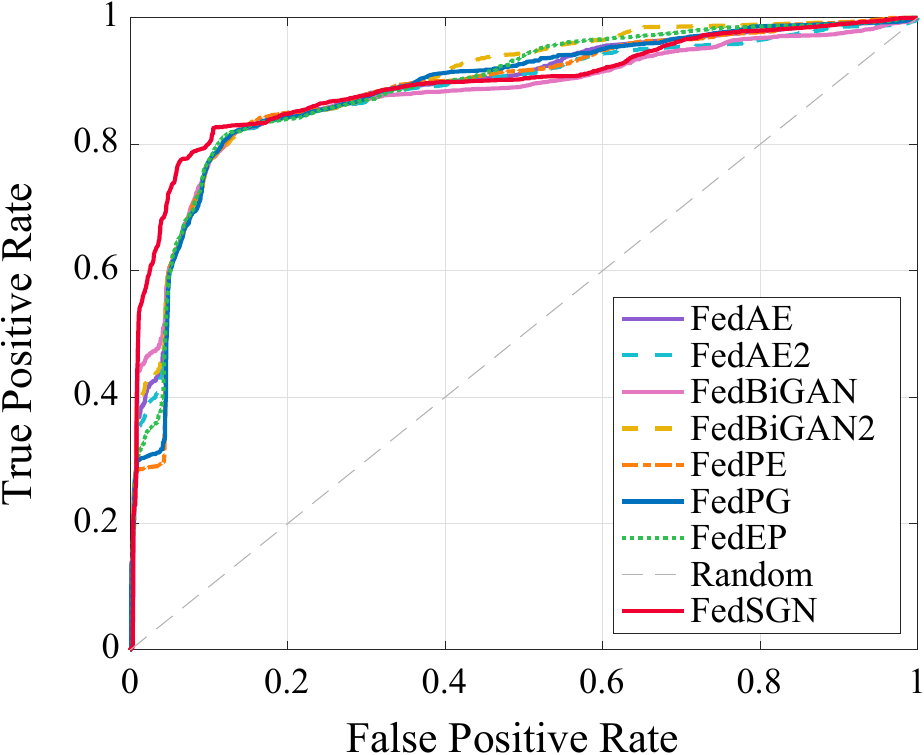}%
		\label{fig:roc_kdd}%
	}\qquad
	\subfigure[TON-IoT]{%
		\includegraphics[width=0.30\textwidth]{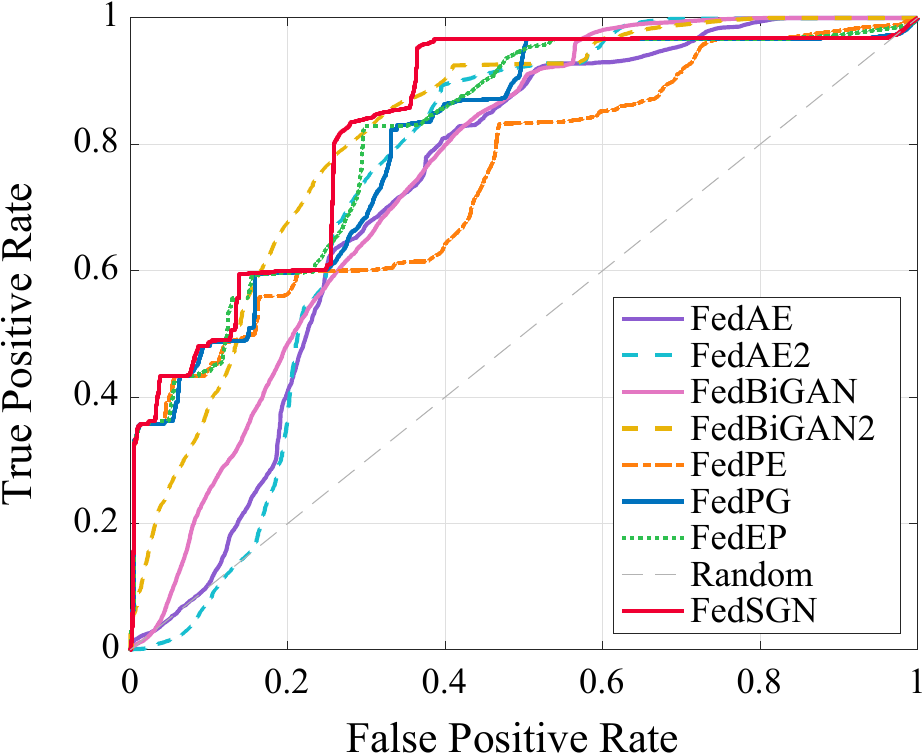}%
		\label{fig:roc_ton}%
	}
%	\vspace{-2mm}
	\caption{ROC curves of the compared methods on the three real-world datasets.}
	\label{fig:ROC}%
\end{figure*}

Furthermore, Fig. \ref{fig:ROC} provides the ROC curves of all methods across the three datasets. On the UNSW-NB15 dataset, FedSGN shows higher true positive rates over intermediate false positive rate intervals. On the NSL-KDD dataset, the advantage is more apparent at low false positive rates, while the curves become closer and cross in subsequent intervals. On the TON-IoT dataset, it also performs favorably in several low and intermediate false positive rate intervals. Overall, these curves complement the ACC and F1 results by showing favorable detection tradeoffs over several threshold ranges.

\subsubsection{Convergence Behavior}
\label{sec:convergence_experiment}

\begin{figure*}[!tbp]
    \centering
    \subfigure[UNSW-NB15]{%
        \includegraphics[width=0.30\textwidth]{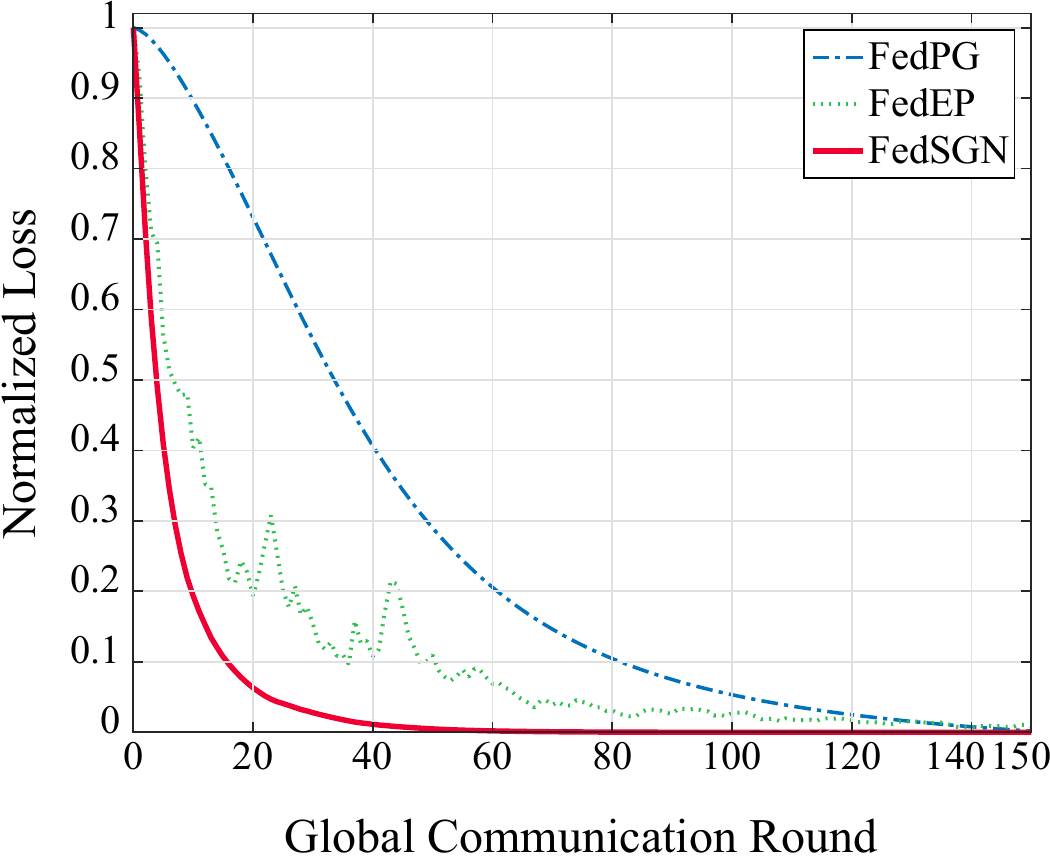}%
        \label{fig:conv_unsw}%
    }\qquad
    \subfigure[NSL-KDD]{%
        \includegraphics[width=0.30\textwidth]{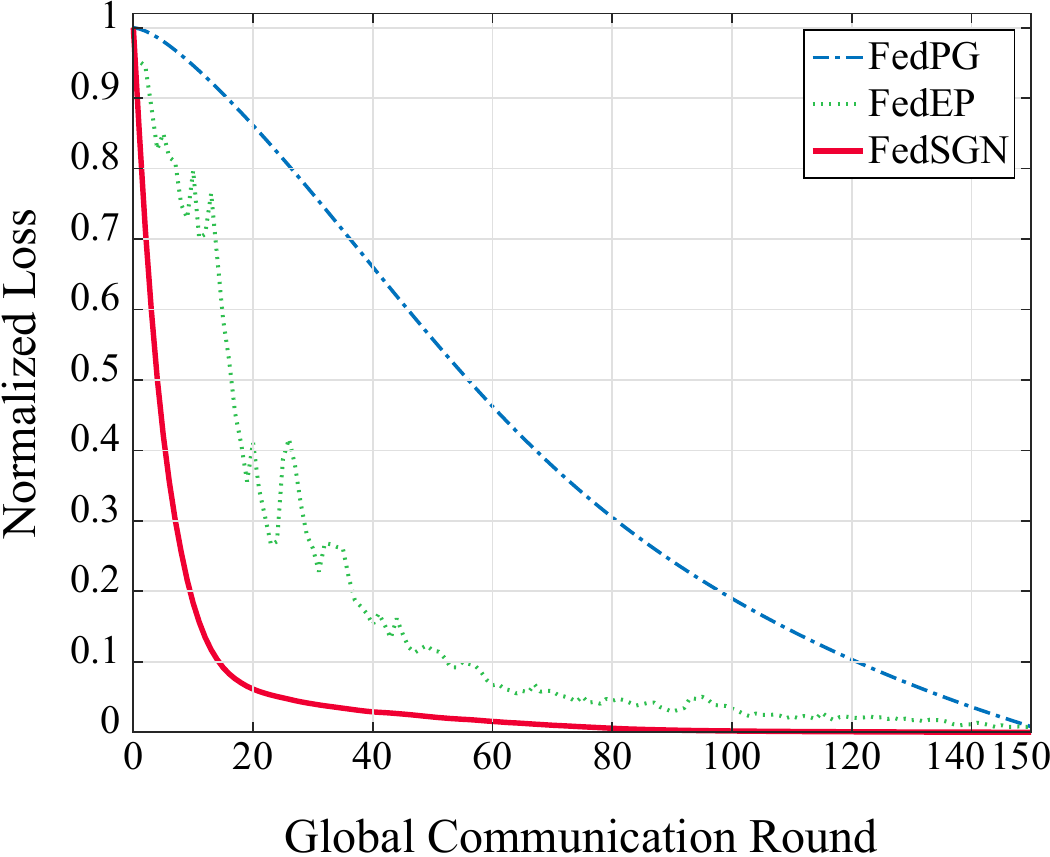}%
        \label{fig:conv_kdd}%
    }\qquad
    \subfigure[TON-IoT]{%
        \includegraphics[width=0.30\textwidth]{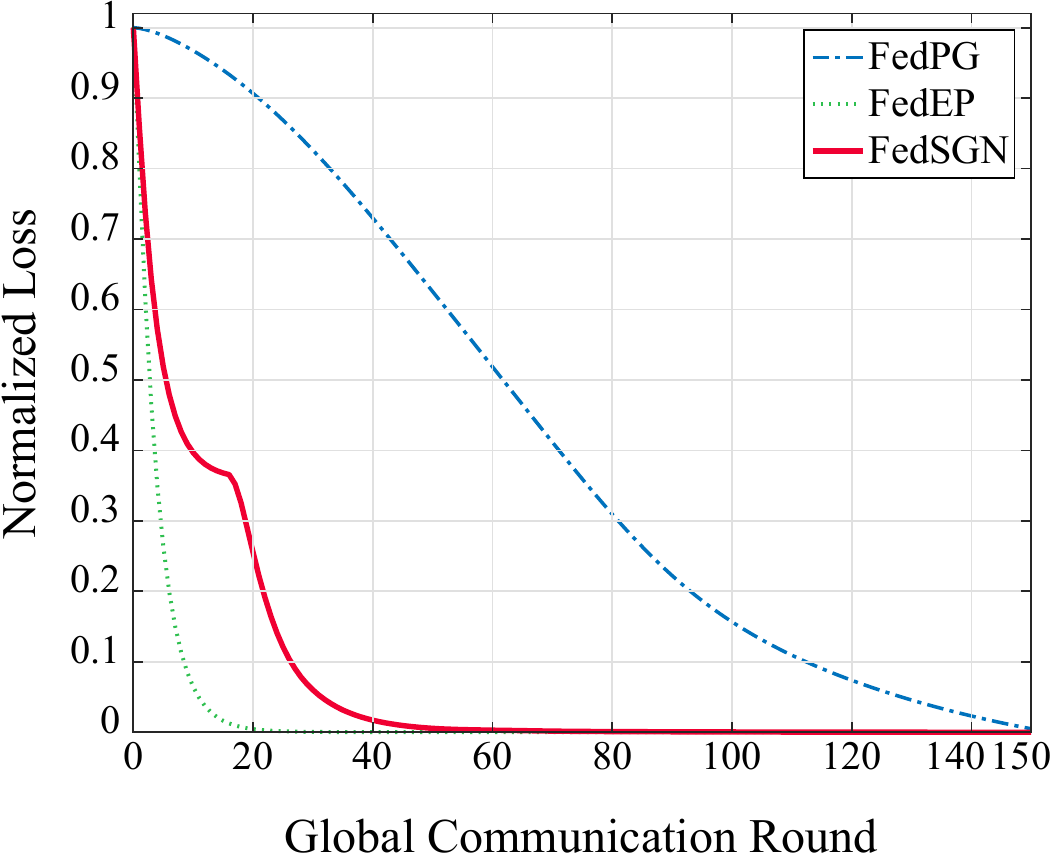}%
        \label{fig:conv_ton}%
    }
    \caption{Normalized convergence curves over 150 global communication rounds.}
    \label{fig:convergence_all}%
\end{figure*}

To investigate convergence behavior, we compare FedPG, FedEP, and FedSGN over 150 global communication rounds. Because the methods optimize different objectives, each loss sequence is independently min-max normalized and causally smoothed using an exponential moving average with a span of seven rounds. The resulting curves illustrate the relative loss reduction of each method.

As illustrated in Fig. \ref{fig:convergence_all}, our FedSGN exhibits a rapid loss decrease during the early rounds, followed by a relatively flat trajectory. This indicates that most observed loss improvement occurs early, with smaller changes in subsequent iterations. Although the loss curve of FedEP stabilizes earlier on the TON-IoT dataset, our FedSGN also shows a rapid loss decline followed by a relatively flat trajectory. In summary, these observations suggest that leveraging second-order information for sparse updates in FedSGN is effective.

\subsubsection{Sparse Feature Analysis}
\label{sec:feature_analysis}

\begin{figure*}[!tbp]
    \centering
    \includegraphics[width=\textwidth]{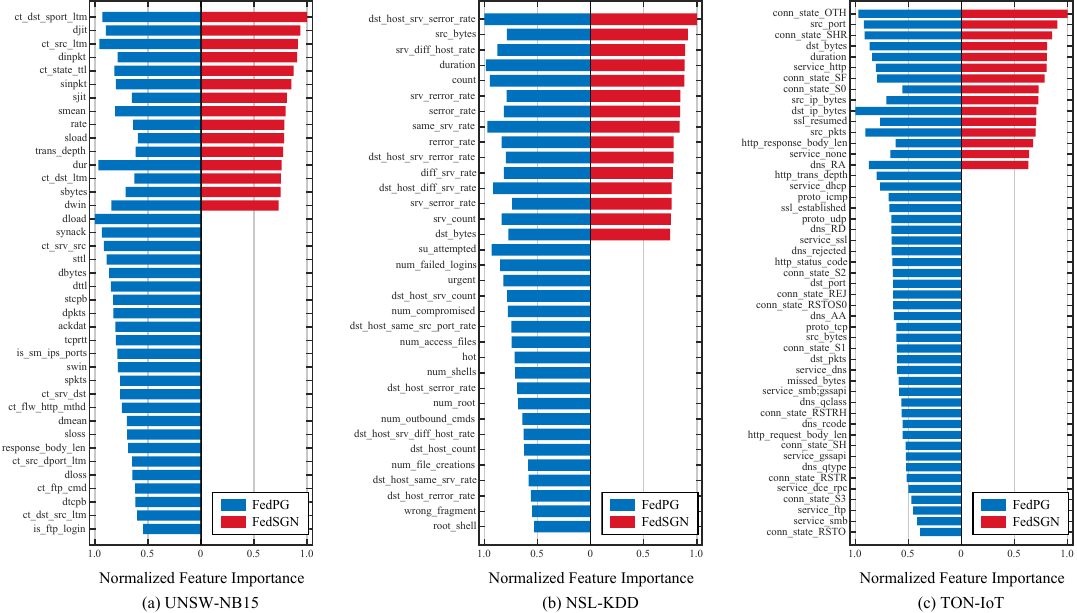}
    \caption{Comparison of feature weights across different datasets.}
    \label{fig:feature_importance}
\end{figure*}

To examine the effects of row-wise sparse constraints on feature selection and model interpretability, we compare FedPG and FedSGN on three real-world datasets. Both methods use identical client partitions and orthogonal initialization, with a subspace dimension of 10, 10 clients, and 30 global communication rounds. Here, feature importance is measured by the $\ell_2$-norm of each row of the learned subspace matrix and normalized by the maximum row norm of each method. 

As shown in Fig. \ref{fig:feature_importance}, blue denotes FedPG and red denotes FedSGN, with FedPG mirrored to the left for comparison. It can be observed that FedPG assigns nonzero weights to all 39, 34, and 49 input features of the UNSW-NB15, NSL-KDD, and TON-IoT datasets, respectively. In contrast, FedSGN retains only 15 feature rows on each dataset and sets all remaining rows to zero. By zeroing unselected rows, FedSGN restricts the global subspace representation to an explicit subset of original features instead of merely shrinking their weights. The retained rows and their relative weights map directly to original data fields, making the features used by the model clearer. These results demonstrate how the $\ell_{2,0}$-norm constraint controls the feature set and makes the learned subspace representation more interpretable.

\subsection{Ablation Studies}
\label{sec:ablation}

In this part, we conduct ablation studies of our proposed FedSGN on the UNSW-NB15 dataset. The experiment uses 10 clients, a 10-dimensional subspace, and 20 global communication rounds, with the sparsity level set to 15. All variants follow the same data preprocessing pipeline and evaluation criteria. The following three variants  are considered
\begin{itemize}
    \item Case I: Replace the Newton direction with the negative gradient direction.
    \item Case II: Remove the $\ell_{2,0}$-norm constraint.
    \item Case III: Remove the Grassmann projection.
\end{itemize}

As listed in Table \ref{tab:ablation_unsw}, our FedSGN  achieves the highest ACC, F1, and AUC among the evaluated configurations. In Case I, the resulting detection performance is close to the complete model, which indicates that replacing the Newton direction with the negative gradient direction alters the optimization strategy, but it does not change the underlying mathematical form of the model itself. In fact, the primary purpose of introducing the Newton direction is to accelerate model convergence while maintaining a high level of detection performance. In Case II, removing the row-wise sparse constraints leads to a decline in ACC and F1 by approximately 7.3\% and 5.2\%, respectively. This suggests that selecting 15 feature rows enables a more compact representation without compromising detection performance. Furthermore, the retained feature rows correspond directly to the original data features and facilitate transmission via sparse matrices. In Case III, removing the Grassmann projection results in a decrease in ACC and F1 by 4.83\% and 2.84\%, respectively, which confirms the effectiveness of this projection step. It should be noted that this variant removes only the Grassmann projection while retaining the orthogonality penalty term used during the training process.

\begin{table}[!t]
    \centering
    \caption{Component ablation results on the UNSW-NB15 dataset.}
    \label{tab:ablation_unsw}
    \renewcommand\arraystretch{1.12}
    \setlength{\tabcolsep}{4pt}
    \resizebox{\columnwidth}{!}{%
    \begin{tabular}{lcccccc}
        \toprule
        Variants & Sparsity & Newton & Grassmann & ACC (\%) & F1 (\%) & AUC (\%) \\
        \midrule
        Case I & $\checkmark$ & $\times$ & $\checkmark$ & 87.25 & 91.29 & 83.26 \\
        Case II & $\times$ & $\checkmark$ & $\checkmark$ & 80.59 & 86.54 & 81.17 \\
        
        Case III & $\checkmark$ & $\checkmark$ & $\times$ & 83.09 & 88.86 & 81.96 \\
        \rowcolor{gray!20} FedSGN & $\checkmark$ & $\checkmark$ & $\checkmark$ & \textbf{87.92} & \textbf{91.70} & \textbf{83.60} \\
        \bottomrule
    \end{tabular}}
\end{table}

\begin{figure}[!tbp]
    \centering
    \includegraphics[width=0.85\columnwidth]{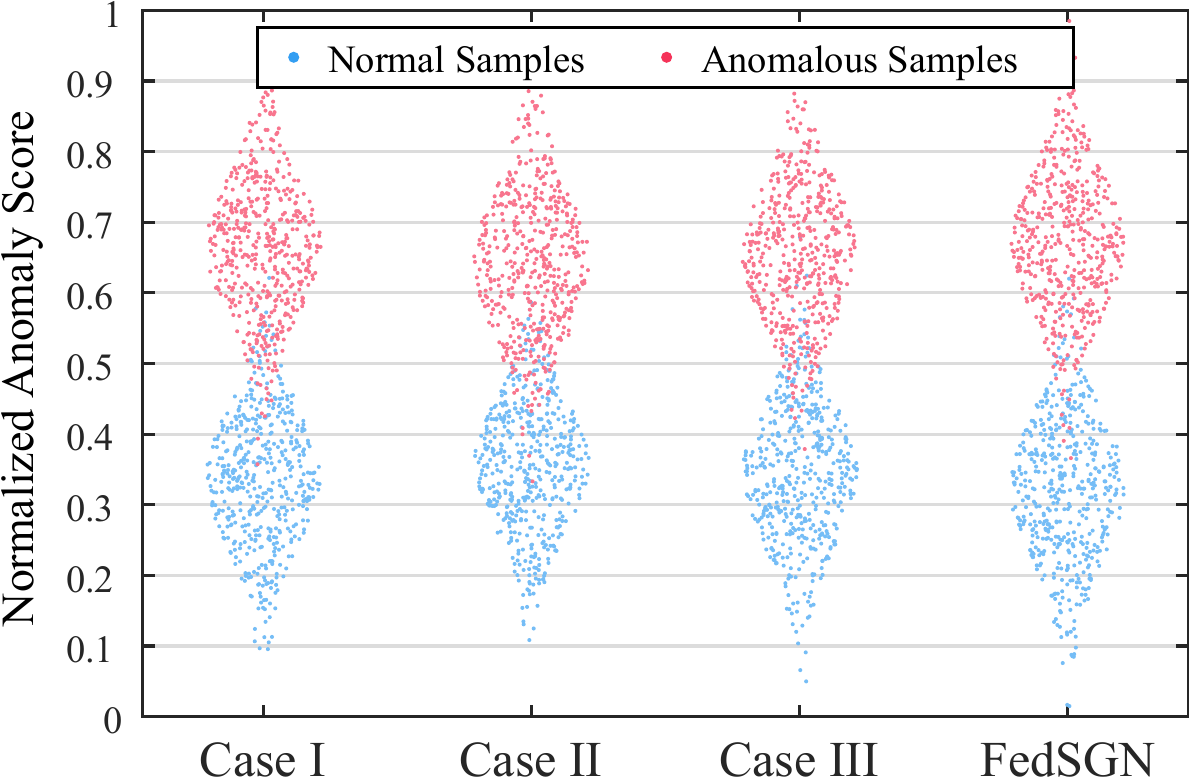}
    \caption{Visualization of normal and anomaly score distributions.}
    \label{fig:ablation_distribution}
\end{figure}

Fig. \ref{fig:ablation_distribution} presents the normal and anomaly score distributions of our FedSGN and its three variants on the UNSW-NB15 dataset, with blue and red denoting normal and anomalous samples, respectively. Evidently, FedSGN places most normal samples in the lower score region and anomalous samples in the higher region, showing clear separation between the two classes. Note that the score distributions for Case II and Case III exhibit significant overlap in the middle range, a finding that corroborates the results presented in Table \ref{tab:ablation_unsw}. Together, these observations support the roles of row-wise sparse constraints and the Grassmann projection in learning a discriminative normal data  subspace.

\subsection{Discussion}
\label{sec:efficiency}

\subsubsection{Communication Efficiency}

\begin{figure}[!tbp]
    \centering
    \includegraphics[width=0.85\columnwidth]{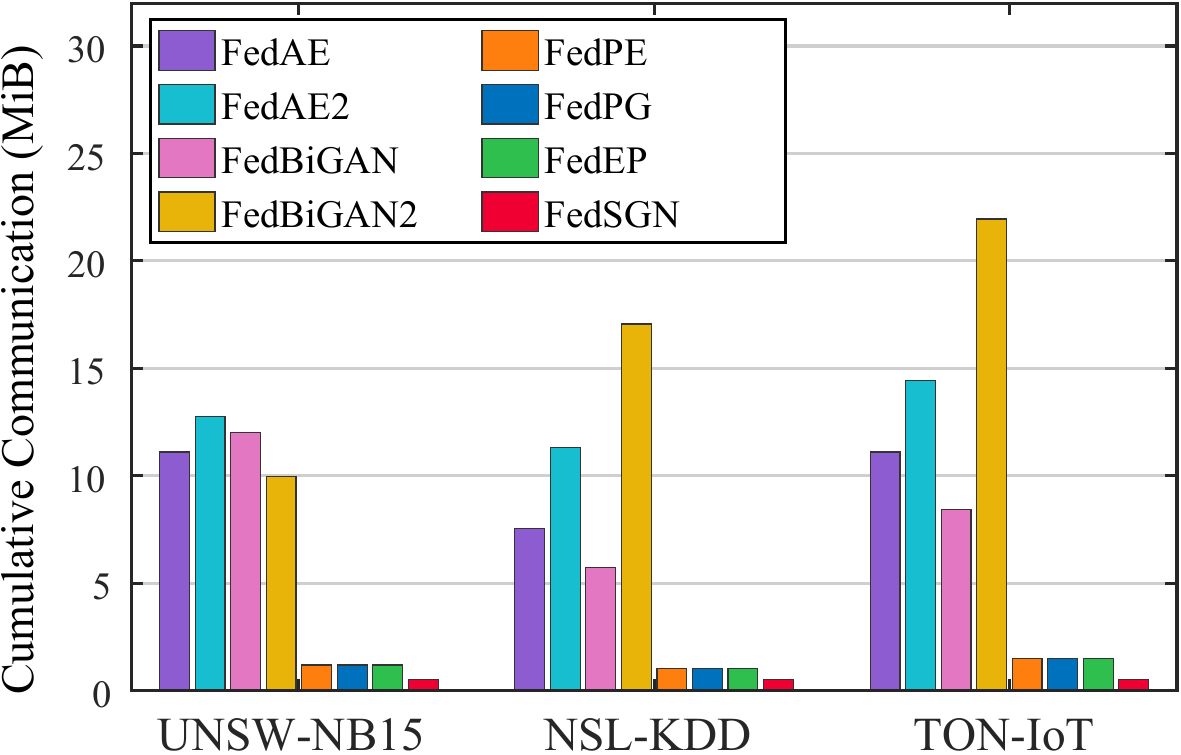}
    \caption{Model transmission volume on the three evaluated datasets.}
    \label{fig:communication_efficiency}
\end{figure}

In the communication comparison experiments, we set up 20 clients and 20 iterations, with each client performing one model representation transmission during the two-way communication phase in every iteration. For all four federated subspace methods, the subspace dimension is set to 10. FedSGN transmits its global projection matrix in sparse form, retaining 15 nonzero feature rows. As shown in Fig. \ref{fig:communication_efficiency}, our FedSGN has a smaller model transmission volume than the compared methods under this common exchange budget. This result illustrates how explicit feature selection produces a more compact model representation.

\subsubsection{Inference Efficiency}

\begin{figure}[!tbp]
    \centering
    \includegraphics[width=0.85\columnwidth]{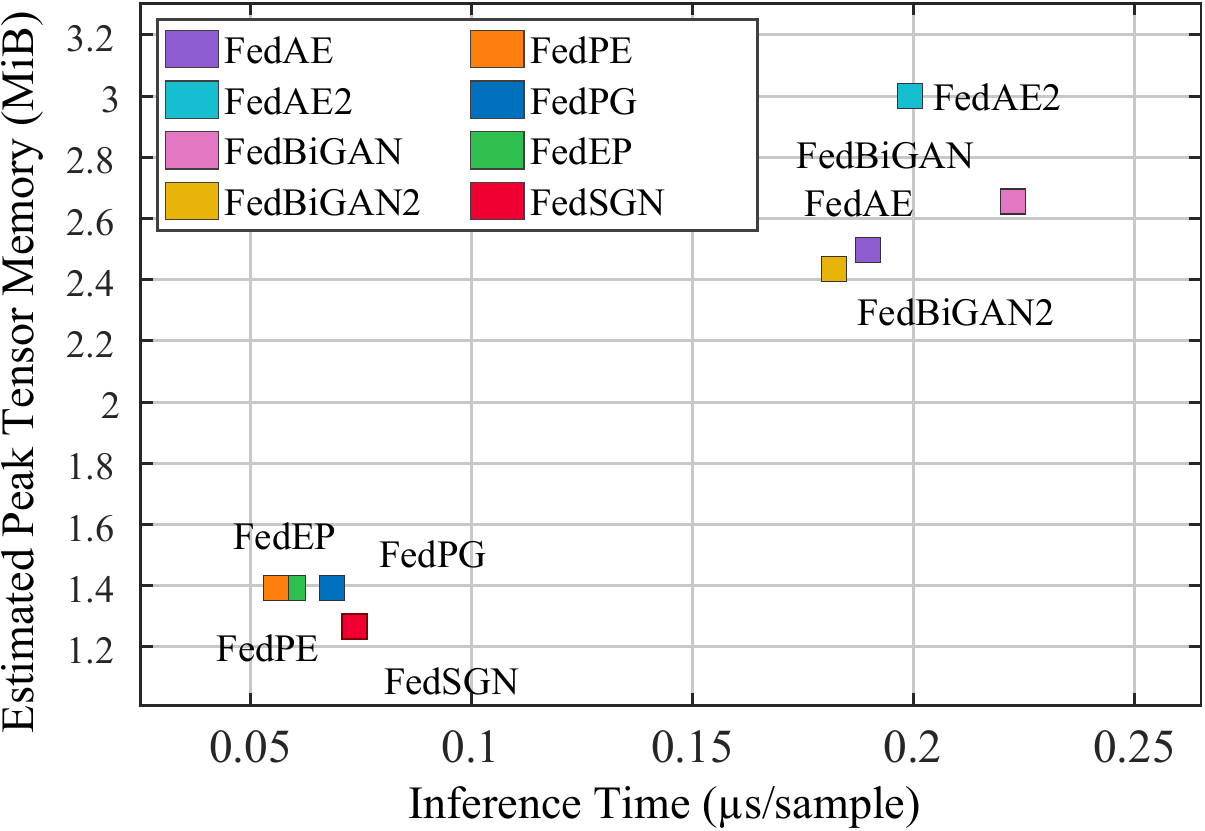}
    \caption{Estimated peak tensor memory versus inference time.}
    \label{fig:inference_efficiency}
\end{figure}

For all four subspace methods, the subspace dimension is set to 10, and a consistent batch size of 4096 normalized samples is used. As shown in Fig. \ref{fig:inference_efficiency}, our FedSGN achieves the lowest estimated tensor storage and shorter inference time compared to neural baseline methods. Compared with the other subspace methods, FedSGN uses less estimated tensor storage but takes slightly longer.

\section{Conclusion}\label{conclusions}

This paper has proposed a novel federated learning method for IoT anomaly detection. Specifically, our FedSGN imposes an $\ell_{2,0}$-norm constraint on the global projection matrix to learn a shared subspace and explicitly select key features. 
By incorporating an orthogonality penalty and performing Grassmann projection in each iteration, FedSGN maintains an effective subspace representation. Moreover, we developed an augmented Lagrangian algorithm with parallel client updates and semismooth Newton optimization for the global sparse update. Numerous experiments have demonstrated favorable overall detection performance of the proposed method when compared against benchmark FedPG and FedEP.

In the future, we focus on the following three research directions. First, developing more efficient optimization algorithms to enhance inference efficiency. Second, adaptive sparsity mechanisms will be designed to improve model generalization. Third, extending the proposed method to neural networks is of importance for practical applications.

\bibliographystyle{IEEEtran}

\bibliography{mybibfile}

\begin{thebibliography}{10}
\providecommand{\url}[1]{#1}
\csname url@samestyle\endcsname
\providecommand{\newblock}{\relax}
\providecommand{\bibinfo}[2]{#2}
\providecommand{\BIBentrySTDinterwordspacing}{\spaceskip=0pt\relax}
\providecommand{\BIBentryALTinterwordstretchfactor}{4}
\providecommand{\BIBentryALTinterwordspacing}{\spaceskip=\fontdimen2\font plus
\BIBentryALTinterwordstretchfactor\fontdimen3\font minus
  \fontdimen4\font\relax}
\providecommand{\BIBforeignlanguage}[2]{{%
\expandafter\ifx\csname l@#1\endcsname\relax
\typeout{** WARNING: IEEEtran.bst: No hyphenation pattern has been}%
\typeout{** loaded for the language `#1'. Using the pattern for}%
\typeout{** the default language instead.}%
\else
\language=\csname l@#1\endcsname
\fi
#2}}
\providecommand{\BIBdecl}{\relax}
\BIBdecl

\bibitem{aouedi2024survey}
O.~Aouedi, T.-H. Vu, A.~Sacco, D.~C. Nguyen, K.~Piamrat, G.~Marchetto, and
  Q.-V. Pham, ``A survey on intelligent {I}nternet of things: Applications,
  security, privacy, and future directions,'' \emph{IEEE Communications Surveys
  \& Tutorials}, vol.~27, no.~2, pp. 1238--1292, 2025.

\bibitem{rejeb2022big}
A.~Rejeb, K.~Rejeb, S.~Simske, H.~Treiblmaier, and S.~Zailani, ``The big
  picture on the {I}nternet of things and the smart city: A review of what we
  know and what we need to know,'' \emph{Internet of Things}, vol.~19, p.
  100565, 2022.

\bibitem{zhou2017security}
J.~Zhou, Z.~Cao, X.~Dong, and A.~V. Vasilakos, ``Security and privacy for
  cloud-based {IoT}: Challenges,'' \emph{IEEE Communications Magazine},
  vol.~55, no.~1, pp. 26--33, 2017.

\bibitem{tang2022computational}
S.~Tang, L.~Chen, K.~He, J.~Xia, L.~Fan, and A.~Nallanathan, ``Computational
  intelligence and deep learning for next-generation edge-enabled industrial
  {IoT},'' \emph{IEEE Transactions on Network Science and Engineering},
  vol.~10, no.~5, pp. 2881--2893, 2023.

\bibitem{tu2025distributed}
J.~Tu, L.~Yang, and J.~Cao, ``Distributed machine learning in edge computing:
  Challenges, solutions and future directions,'' \emph{ACM Computing Surveys},
  vol.~57, no.~5, pp. 1--37, 2025.

\bibitem{liu2024deafl}
S.~Liu, Y.~Yu, Y.~Zong, P.~L. Yeoh, L.~Guo, B.~Vucetic, T.~Q. Duong, and Y.~Li,
  ``Delay and energy-efficient asynchronous federated learning for intrusion
  detection in heterogeneous industrial internet of things,'' \emph{IEEE
  Internet of Things Journal}, vol.~11, no.~8, pp. 14\,739--14\,754, 2024.

\bibitem{adhikari2024recent}
D.~Adhikari, W.~Jiang, J.~Zhan, D.~B. Rawat, and A.~Bhattarai, ``Recent
  advances in anomaly detection in internet of things: Status, challenges, and
  perspectives,'' \emph{Computer Science Review}, vol.~54, p. 100665, 2024.

\bibitem{kounoudes2020mapping}
A.~D. Kounoudes and G.~M. Kapitsaki, ``A mapping of {IoT} user-centric privacy
  preserving approaches to the {GDPR},'' \emph{Internet of Things}, vol.~11, p.
  100179, 2020.

\bibitem{li2025flow}
Z.~Li, W.~Yao, J.~Luo, and Z.~Huang, ``Flow-based {IoT} intrusion detection via
  improved generative federated distillation learning,'' \emph{IEEE Internet of
  Things Journal}, vol.~12, no.~10, pp. 14\,797--14\,811, 2025.

\bibitem{mcmahan2017communication}
B.~McMahan, E.~Moore, D.~Ramage, S.~Hampson, and B.~A.~y. Arcas,
  ``Communication-efficient learning of deep networks from decentralized
  data,'' in \emph{Artificial Intelligence and Statistics}.\hskip 1em plus
  0.5em minus 0.4em\relax PMLR, 2017, pp. 1273--1282.

\bibitem{nguyen2021federated}
D.~C. Nguyen, M.~Ding, P.~N. Pathirana, A.~Seneviratne, J.~Li, and H.~V. Poor,
  ``Federated learning for {I}nternet of things: A comprehensive survey,''
  \emph{IEEE Communications Surveys \& Tutorials}, vol.~23, no.~3, pp.
  1622--1658, 2021.

\bibitem{li2020federated}
T.~Li, A.~K. Sahu, M.~Zaheer, M.~Sanjabi, A.~Talwalkar, and V.~Smith,
  ``Federated optimization in heterogeneous networks,'' \emph{Proceedings of
  Machine Learning and Systems}, vol.~2, pp. 429--450, 2020.

\bibitem{lin2025fidscl}
Q.~Lin, W.~Liu, S.~Zheng, J.~Ji, K.-C. Wong, J.~Li, and C.~A. Coello~Coello,
  ``Federated intrusion detection system with cost-sensitive learning for
  internet of things,'' \emph{IEEE Internet of Things Journal}, vol.~12,
  no.~19, pp. 39\,677--39\,688, 2025.

\bibitem{hu2025efficient}
T.~Hu, Q.~Chen, Y.~Hu, S.~Hou, H.~Yan, P.~Yi, and Z.~Cui, ``Efficient and
  privacy-preserving network intrusion detection based on federated learning in
  {SDN}-enabled {IIoT} network,'' \emph{IEEE Internet of Things Journal},
  vol.~12, no.~20, pp. 41\,904--41\,923, 2025.

\bibitem{abusitta2023deep}
A.~Abusitta, G.~H. de~Carvalho, O.~A. Wahab, T.~Halabi, B.~C. Fung, and
  S.~Al~Mamoori, ``Deep learning-enabled anomaly detection for {IoT} systems,''
  \emph{Internet of Things}, vol.~21, p. 100656, 2023.

\bibitem{huang2025deep}
H.~Huang, P.~Wang, J.~Pei, J.~Wang, S.~Alexanian, and D.~Niyato, ``Deep
  learning advancements in anomaly detection: A comprehensive survey,''
  \emph{IEEE Internet of Things Journal}, vol.~12, no.~21, pp.
  44\,318--44\,342, 2025.

\bibitem{ieracitano2020novel}
C.~Ieracitano, A.~Adeel, F.~C. Morabito, and A.~Hussain, ``A novel statistical
  analysis and autoencoder driven intelligent intrusion detection approach,''
  \emph{Neurocomputing}, vol. 387, pp. 51--62, 2020.

\bibitem{ruff2021unifying}
L.~Ruff, J.~R. Kauffmann, R.~A. Vandermeulen, G.~Montavon, W.~Samek, M.~Kloft,
  T.~G. Dietterich, and K.-R. M{\"u}ller, ``A unifying review of deep and
  shallow anomaly detection,'' \emph{Proceedings of the IEEE}, vol. 109, no.~5,
  pp. 756--795, 2021.

\bibitem{nguyen2024federated}
T.-A. Nguyen, L.~T. Le, T.~D. Nguyen, W.~Bao, S.~Seneviratne, C.~S. Hong, and
  N.~H. Tran, ``Federated {PCA} on {G}rassmann manifold for {IoT} anomaly
  detection,'' \emph{IEEE/ACM Transactions on Networking}, vol.~32, no.~5, pp.
  4456--4471, 2024.

\bibitem{donahue2017adversarial}
J.~Donahue, P.~Kr{\"a}henb{\"u}hl, and T.~Darrell, ``Adversarial feature
  learning,'' in \emph{Proc. Int. Conf. Learn. Represent.}, 2017.

\bibitem{zhou2024reconstructed}
X.~Zhou, J.~Wu, W.~Liang, K.~I.-K. Wang, Z.~Yan, L.~T. Yang, and Q.~Jin,
  ``Reconstructed graph neural network with knowledge distillation for
  lightweight anomaly detection,'' \emph{IEEE Transactions on Neural Networks
  and Learning Systems}, vol.~35, no.~9, pp. 11\,817--11\,828, 2024.

\bibitem{kalakoti2026synthetic}
R.~Kalakoti, H.~Bahsi, and S.~N{\~o}mm, ``Synthetic data-driven explainability
  for federated learning-based intrusion detection system,'' \emph{IEEE
  Internet of Things Journal}, vol.~13, no.~4, pp. 5918--5944, 2026.

\bibitem{hoang2018pca}
D.~H. Hoang and H.~D. Nguyen, ``A {PCA}-based method for {IoT} network traffic
  anomaly detection,'' in \emph{2018 20th International Conference on Advanced
  Communication Technology (ICACT)}.\hskip 1em plus 0.5em minus 0.4em\relax
  IEEE, 2018, pp. 381--386.

\bibitem{liu2025similarity}
Z.~Liu, H.~Lv, X.~Liu, C.~Ma, F.~Wu, L.~Liu, and L.~Cui, ``Similarity and
  diversity: {PCA}-based contribution evaluation in federated learning,''
  \emph{IEEE Internet of Things Journal}, vol.~12, no.~12, pp.
  20\,393--20\,405, 2025.

\bibitem{carter2022fast}
J.~Carter, S.~Mancoridis, and E.~Galinkin, ``Fast, lightweight {IoT} anomaly
  detection using feature pruning and {PCA},'' in \emph{Proceedings of the 37th
  ACM/SIGAPP Symposium on Applied Computing}, 2022, pp. 133--138.

\bibitem{boyd2011distributed}
S.~Boyd, N.~Parikh, E.~Chu, B.~Peleato, and J.~Eckstein, ``Distributed
  optimization and statistical learning via the alternating direction method of
  multipliers,'' \emph{Foundations and Trends in Machine Learning}, vol.~3,
  no.~1, pp. 1--122, 2011.

\bibitem{xiu2026fedep}
X.~Xiu, C.~Huang, W.~Zhang, and W.~Liu, ``Efficient personalized federated
  {PCA} with manifold optimization for {IoT} anomaly detection,'' \emph{IEEE
  Internet of Things Journal}, vol.~13, no.~16, pp. 37\,252--37\,262, 2026.

\bibitem{liu2026bi}
J.~Liu, X.~Ju, X.~Xiu, and W.~Liu, ``Bi-level unsupervised feature selection
  with {$\ell_{2,0}$}-norm,'' \emph{Pattern Recognition}, p. 114687, 2026.

\bibitem{li2026enhancing}
X.~Li, C.~Huang, A.~Yang, and X.~Xiu, ``Enhancing unsupervised feature
  selection via double sparsity constrained optimization,''
  \emph{Knowledge-Based Systems}, p. 116924, 2026.

\bibitem{ghubaish2024lemda}
A.~Ghubaish, Z.~Yang, A.~Erbad, and R.~Jain, ``{LEMDA}: A novel feature
  engineering method for intrusion detection in {IoT} systems,'' \emph{IEEE
  Internet of Things Journal}, vol.~11, no.~8, pp. 13\,247--13\,256, 2024.

\bibitem{zou2006sparse}
H.~Zou, T.~Hastie, and R.~Tibshirani, ``Sparse principal component analysis,''
  \emph{Journal of Computational and Graphical Statistics}, vol.~15, no.~2, pp.
  265--286, 2006.

\bibitem{journee2010generalized}
M.~Journ{\'e}e, Y.~Nesterov, P.~Richt{\'a}rik, and R.~Sepulchre, ``Generalized
  power method for sparse principal component analysis,'' \emph{Journal of
  Machine Learning Research}, vol.~11, no.~15, pp. 517--553, 2010.

\bibitem{dong2025unsupervised}
X.~Dong, F.~Nie, L.~Tian, R.~Wang, and X.~Li, ``Unsupervised discriminative
  feature selection with $\ell_{2,0}$-norm constrained sparse projection,''
  \emph{IEEE Transactions on Pattern Analysis and Machine Intelligence},
  vol.~47, no.~10, pp. 8321--8335, 2025.

\bibitem{zhu2025joint}
Y.~Zhu, X.~Xiu, W.~Liu, and C.~Yin, ``Joint sparse subspace clustering via fast
  $\ell_{2,0}$-norm constrained optimization,'' \emph{Expert Systems with
  Applications}, vol. 265, p. 125845, 2025.

\bibitem{xiu2022sparsity}
X.~Xiu, Z.~Miao, and W.~Liu, ``A sparsity-aware fault diagnosis framework
  focusing on accurate isolation,'' \emph{IEEE Transactions on Industrial
  Informatics}, vol.~19, no.~2, pp. 1356--1365, 2023.

\bibitem{chen2020alternating}
S.~Chen, S.~Ma, L.~Xue, and H.~Zou, ``An alternating manifold proximal gradient
  method for sparse principal component analysis and sparse canonical
  correlation analysis,'' \emph{INFORMS Journal on Optimization}, vol.~2,
  no.~3, pp. 192--208, 2020.

\bibitem{li2025riemannian}
J.~Li, S.~Ma, and T.~Srivastava, ``A {R}iemannian alternating direction method
  of multipliers,'' \emph{Mathematics of Operations Research}, vol.~50, no.~4,
  pp. 3222--3242, 2025.

\bibitem{grammenos2020federated}
A.~Grammenos, R.~Mendoza-Smith, J.~Crowcroft, and C.~Mascolo, ``Federated
  principal component analysis,'' in \emph{Advances in Neural Information
  Processing Systems}, vol.~33, 2020, pp. 6453--6464.

\bibitem{wang2023high}
K.~Wang and Z.~Song, ``High-dimensional cross-plant process monitoring with
  data privacy: A federated hierarchical sparse {PCA} approach,'' \emph{IEEE
  Transactions on Industrial Informatics}, vol.~20, no.~3, pp. 4385--4396,
  2024.

\bibitem{qu2024distributed}
W.~Qu, H.~Chen, X.~Xiu, and W.~Liu, ``Distributed sparsity constrained
  optimization over the {Stiefel} manifold,'' \emph{Neurocomputing}, vol. 602,
  p. 128267, 2024.

\bibitem{xiao2018regularized}
X.~Xiao, Y.~Li, Z.~Wen, and L.~Zhang, ``A regularized semi-smooth {N}ewton
  method with projection steps for composite convex programs,'' \emph{Journal
  of Scientific Computing}, vol.~76, no.~1, pp. 364--389, 2018.

\bibitem{zhou2021global}
S.~Zhou, N.~Xiu, and H.-D. Qi, ``Global and quadratic convergence of {Newton}
  hard-thresholding pursuit,'' \emph{Journal of Machine Learning Research},
  vol.~22, no.~12, pp. 1--45, 2021.

\bibitem{hinton2006reducing}
G.~E. Hinton and R.~R. Salakhutdinov, ``Reducing the dimensionality of data
  with neural networks,'' \emph{Science}, vol. 313, no. 5786, pp. 504--507,
  2006.

\bibitem{moustafa2015unsw}
N.~Moustafa and J.~Slay, ``{UNSW-NB15}: A comprehensive data set for network
  intrusion detection systems ({UNSW-NB15} network data set),'' in \emph{2015
  Military Communications and Information Systems Conference (MilCIS)}.\hskip
  1em plus 0.5em minus 0.4em\relax IEEE, 2015, pp. 1--6.

\bibitem{tavallaee2009detailed}
M.~Tavallaee, E.~Bagheri, W.~Lu, and A.~A. Ghorbani, ``A detailed analysis of
  the {KDD} {CUP} 99 data set,'' in \emph{2009 IEEE Symposium on Computational
  Intelligence for Security and Defense Applications}.\hskip 1em plus 0.5em
  minus 0.4em\relax IEEE, 2009, pp. 1--6.

\bibitem{booij2021ton_iot}
T.~M. Booij, I.~Chiscop, E.~Meeuwissen, N.~Moustafa, and F.~T. Den~Hartog,
  ``{ToN}\_{IoT}: The role of heterogeneity and the need for standardization of
  features and attack types in {IoT} network intrusion data sets,'' \emph{IEEE
  Internet of Things Journal}, vol.~9, no.~1, pp. 485--496, 2022.

\end{thebibliography}

\end{document}